\documentclass{article}

\usepackage[preprint]{neurips_2026}

\usepackage[utf8]{inputenc} 
\usepackage[T1]{fontenc}    
\usepackage{hyperref}       
\usepackage{url}            
\usepackage{booktabs}       
\usepackage{amsfonts}       
\usepackage{nicefrac}       
\usepackage{microtype}      
\usepackage{xcolor}         

\usepackage{enumitem}
\usepackage{amsmath}
\usepackage{amssymb}
\usepackage{amsthm}
\usepackage{multirow}
\usepackage{booktabs}
\usepackage{siunitx}
\usepackage{graphicx}
\usepackage{xcolor}
\usepackage[table]{xcolor}
\usepackage{colortbl}
\usepackage{caption}
\usepackage{adjustbox}
\usepackage{pifont}

\definecolor{mygray}{gray}{0.9}

\title{Only Say What You Know: Calibration-Aware Generation for Long-Form Factuality}

\author{
Wen Luo\thanks{Work done during internship at MSRA.}\textsuperscript{\rm \,\, $\heartsuit$},
Guangyue Peng\textsuperscript{\rm $\heartsuit$},
Liang Wang\textsuperscript{\ding{171}},
Nan Yang\textsuperscript{\ding{171}},
Wei Li\textsuperscript{\rm $\heartsuit$},
\\
\textbf{
Yuhan Song\textsuperscript{\rm $\heartsuit$},
Shaohang Wei\textsuperscript{\rm $\heartsuit$},
Feifan Song\textsuperscript{\rm $\heartsuit$},
Furu Wei\textsuperscript{\ding{171}},
Houfeng Wang\thanks{Corresponding author}\textsuperscript{\rm \,\, $\heartsuit$}}\\
\textsuperscript{\rm $\heartsuit$} State Key Laboratory of Multimedia Information Processing,\\
School of Computer Science, Peking University\\
\textsuperscript{\ding{171}} Microsoft Research Asia
}

\begin{document}

\maketitle

\begin{abstract}

Large Reasoning Models achieve strong performance on complex tasks but remain prone to hallucinations, particularly in long-form generation where errors compound across reasoning steps. 
Existing approaches to improving factuality, including abstention and factuality-driven optimization, follow a \emph{coupled exploration-commitment} paradigm, in which intermediate reasoning is unconditionally propagated to the final output, limiting fine-grained control over information selection and integration.
In this paper, we propose an \textbf{Exploration-Commitment Decoupling} paradigm that disentangles knowledge exploration from final commitment, enabling models to explore with awareness while answering cautiously.
We instantiate the paradigm with \textbf{Calibration-Aware Generation (CAG)}, a framework that equips models with end-to-end, calibration-aware generation capabilities, by augmenting intermediate reasoning with calibrated reliability estimates and prioritizing reliable content in final outputs.
Across five long-form factuality benchmarks and multiple model families, CAG improves factuality by up to 13\%, while reducing decoding time by up to 37\%. 
Overall, our work highlights decoupling as a principled approach for more reliable long-form generation, offering directions for trustworthy and self-aware generative systems.

\end{abstract}

\section{Introduction}
\label{sec:intro}

Large Reasoning Models (LRMs), such as DeepSeek-R1 \citep{deepseekai2025deepseekv3technicalreport} and GPT-5 \citep{singh2025openaigpt5card}, have achieved impressive performance across diverse domains, including mathematics and code generation.
However, despite their advanced reasoning capabilities, LRMs are still not fully reliable: they frequently produce \emph{hallucinations}—plausible but factually incorrect outputs. This limitation poses a major obstacle to the real-world deployment of LRMs, where reliability and trustworthiness are essential \citep{ye-etal-2025-toolhop,DBLP:conf/acl/HuCSCWLZ0Y25}. 
Moreover, reasoning models are often more prone to hallucinations than their non-reasoning counterparts \citep{chen2025learningreasonfactuality}, making the consequences of such errors increasingly severe, particularly in long-form settings.

Recent efforts to improve factuality generally follow two main directions.
One line of work \citep{wu2026mitigatingllmhallucinationbehaviorally,an2025teachingllmsabstainfinegrained,DBLP:conf/acl/YuanJ00XLHT25} encourages models to abstain when they are likely to be incorrect \citep{ren2025knowrlexploringknowledgeablereinforcement,xue-etal-2025-ualign}.
Another line of work directly optimizes factuality through specialized objectives, such as supervised fine-tuning \citep{DBLP:conf/acl/XieZPJMFTWFRFZ25,DBLP:conf/nips/LinGOXLY024} or reinforcement learning with factuality-oriented rewards \citep{chen2025learningreasonfactuality,chen2025traintruthskillsbinary,wu2026mitigatingllmhallucinationbehaviorally}.
Despite their effectiveness, both approaches adhere to a \textbf{coupled exploration–commitment paradigm}, where intermediate reasoning is unconditionally carried into the final output, preventing fine-grained control over information selection and integration.
The limitation is particularly pronounced in long-form generation, where responses contain multiple claims with varying levels of reliability.
Abstention-based methods rely on a coarse answer-or-abstain strategy, reducing errors at the cost of utility and helpfulness \citep{su-etal-2025-ai,DBLP:conf/icml/ChengSLZYLLH0Q24}. 
Factuality-optimized approaches, on the other hand, improve overall output quality but fail to model the reliability of individual reasoning components, often propagating hallucinations into the final answer \citep{chen2025learningreasonfactuality}.

\begin{figure}
\centering
\includegraphics[width=\textwidth]{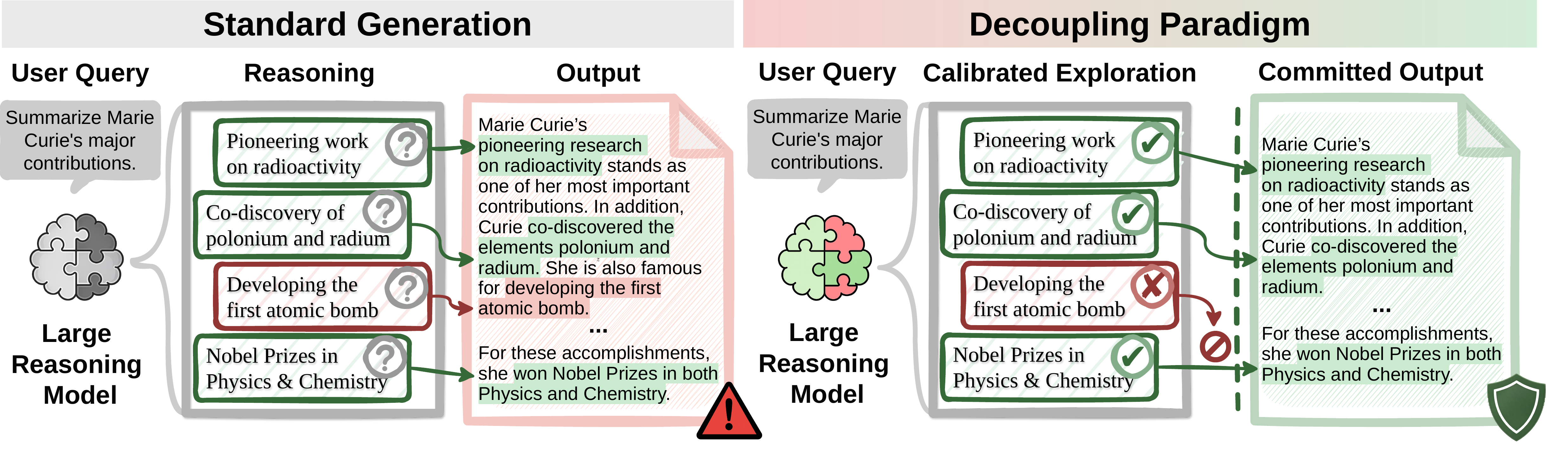}
\caption{Comparison between standard generation (left) and the proposed Exploration–Commitment Decoupling paradigm (right). Standard generation unconditionally propagates intermediate reasoning into the final output, often leading to hallucinations. In contrast, the proposed paradigm introduces calibrated exploration and selective commitment, enabling the model to estimate the reliability of intermediate reasoning and retain only trustworthy information in the final output.}
\label{fig:intro}
\end{figure}

In this paper, we propose an \textbf{Exploration-Commitment Decoupling} paradigm for improving factuality in long-form generation (Figure \ref{fig:intro}).
LRMs typically entertain multiple candidate directions during reasoning, only a subset of which should be reflected in the final answer.
Therefore, the key idea is to disentangle \emph{knowledge exploration} from \emph{final commitment}: 
models should maintain their original exploration process to preserve utility, while remaining conservative in their final outputs to minimize hallucination risks.
To this end, we introduce \textbf{Calibration-Aware Generation (CAG)}, a framework that enables end-to-end, calibration-aware generation by explicitly separating exploration from commitment. 
Specifically, CAG is characterized by two complementary mechanisms:
\textbf{(i) Calibrated exploration: knowing what the model knows.} During reasoning, the model generates well-calibrated intermediate steps, each accompanied by explicit reliability estimates. These signals are learned from fine-grained factuality annotations, encouraging alignment between the model's expressed reliability and actual correctness of its reasoning.
\textbf{(ii) Selective commitment: acting on what the model knows.} After exploration, the model constructs the final answer by selectively retaining reliable content and down-weighting unreliable information, guided by preceding calibrated signals.
In this way, the exploration trajectory (i.e., reasoning) serves as a structured representation that guides commitment (i.e., the final answer), enabling the model to \emph{explore with awareness while committing cautiously}, thereby improving long-form factuality while maintaining end-to-end generation efficiency.
Importantly, the proposed Exploration-Commitment Decoupling paradigm is \emph{method-agnostic}: it does not rely on a specific instantiation or training objective, and can be integrated with a wide range of existing approaches (e.g., factuality-driven optimization or retrieval augmentation), offering a unifying perspective for future method design.

Across 5 benchmarks and 5 LRMs spanning diverse architectures and scales, CAG consistently improves factuality by augmenting the original reasoning structure with calibrated reliability signals and selectively committing to reliable content. It achieves up to a 13\% improvement without compromising overall informativeness.
Notably, CAG generalizes beyond long-form generation to a variety of settings, including knowledge-intensive question answering, chatbot scenarios, and Retrieval-Augmented Generation (RAG), demonstrating its broad applicability.
Furthermore, by suppressing unreliable content, CAG yields additional efficiency gains, reducing decoding time by up to 37\%.

Overall, our key contributions are summarized as follows:

\begin{itemize}[left=0cm]

\item \textbf{(Paradigm)} We propose an \textbf{Exploration-Commitment Decoupling} paradigm that reframes long-form generation in LRMs by disentangling knowledge exploration from final commitment. This paradigm provides a general perspective, enabling fine-grained control over how intermediate reasoning is translated into final outputs.

\item \textbf{(Framework)} We introduce \textbf{Calibration-Aware Generation (CAG)}, a framework that instantiates Exploration-Commitment Decoupling and enables end-to-end, calibration-aware generation in a single pass. CAG integrates \textbf{(i) calibrated exploration}, which equips intermediate reasoning with calibrated factual reliability, and \textbf{(ii) selective commitment}, which constructs final answers by prioritizing reliable content, resulting in improved factuality while preserving informativeness.

\item \textbf{(Improvement)} Extensive experiments across diverse LRMs and tasks show that CAG consistently improves factuality by up to 13\%, generalizes to settings including knowledge-intensive QA, chatbot scenarios, and RAG, and reduces decoding time by up to 37\% by down-weighting unreliable content.

\end{itemize}

\section{Methodology}

In this section, we present the Exploration-Commitment Decoupling paradigm and its instantiation, CAG.
We begin by introducing the problem setup and analyzing the limitations of existing generation paradigms (\S\ref{sec:methodology_setup}). 
We then propose the Exploration-Commitment Decoupling paradigm that disentangles reasoning exploration from answer commitment via step-level reliability estimation.
To realize the paradigm, we introduce a structured supervision framework (\S\ref{sec:methodology_supervision}) that jointly learns calibrated reasoning and selective commitment.
Finally, we further refine the learned policy via distillation (\S\ref{sec:methodology_distillation}), transferring calibration-aware behaviors to enhance robustness and generalization.

\subsection{Problem Setup}
\label{sec:methodology_setup}

We study long-form generation using LRMs.
Formally, given an input query $x$, a reasoning model produces a sequence of $|\mathcal{R}|$ intermediate reasoning steps $r = (r_1, \dots, r_{|\mathcal{R}|})$, followed by a final answer $y$.  
Each reasoning step $r_i$ consists of a sequence of $|\mathcal{A}_i|$ atomic claims 
$(a_{i,1}, \dots, a_{i,|\mathcal{A}_i|})$, where each $a_{i,j}$ represents a minimal, self-contained unit of reasoning \citep{song-etal-2024-veriscore}.
In standard generation, the model produces a reasoning trajectory $r \sim p_\theta(\cdot \mid x)$ and generates the final answer conditioned on the entire reasoning, $y \sim p_\theta(\cdot \mid x, r)$, implicitly treating all reasoning steps equally regardless of their reliability.
However, factuality in long-form generation is inherently fine-grained: different components of the reasoning trajectory may vary significantly in correctness. 
As a result, unreliable reasoning steps may propagate into the final answer, leading to hallucinations.
To address the challenge, we propose \textbf{Exploration-Commitment Decoupling}, a paradigm that explicitly models the reliability of each reasoning step and selectively incorporates reliable content into the final answer.

\subsection{Exploration-Commitment Decoupling}
\label{sec:methodology_decoupling}

We propose the Exploration-Commitment Decoupling paradigm that disentangles knowledge exploration from final commitment in long-form generation.

\paragraph{Formulation}

We first introduce step-level calibration into the reasoning process by associating each reasoning step $r_i$ with a reliability variable $c_i \in \mathcal{C}$, where $\mathcal{C}$ denotes a set of reliability levels. 
The resulting reliability-aware reasoning trajectory is defined as
\begin{equation}
\small
(r, c) = \big((r_1, c_1), \dots, (r_{|\mathcal{R}|}, c_{|\mathcal{R}|})\big).
\end{equation}

The generation process is then decomposed as:
\begin{gather}
\small
(r, c) \sim p_\theta(\cdot \mid x), \\
\tilde{y} \sim p_\theta(\cdot \mid x, r, c),
\end{gather}
where $r$ represents the exploration of candidate reasoning paths, $c$ encodes the model's estimation of reliability over steps, and $\tilde{y}$ denotes the final committed answer. 
Under the proposed formulation, the objective shifts from producing a single monolithic output to learning a calibrated generation policy that (i) jointly generates reasoning steps with well-calibrated reliability estimates $(r, c)$, and (ii) leverages these estimates to construct more reliable final answers $\tilde{y}$.

\paragraph{Calibrated Exploration}
We augment each reasoning step with an explicit reliability variable, yielding a structured representation $(r_i, c_i)$.
\textbf{Importantly, the reasoning itself remains unchanged:} the model is still encouraged to explore as usual.
Instead, the reliability variables, learned from factuality annotations, serve as assessment signals that reflect the model's estimation of each step's factual correctness.
The proposed design enables the model to jointly represent \emph{what it thinks} and \emph{how reliable it is}, providing a fine-grained and interpretable structure over the reasoning trajectory.

\paragraph{Selective Commitment}
Given the augmented reasoning $(r, c)$, the final answer is constructed via selective commitment. 
Specifically, the model learns to prioritize trustworthy reasoning content while down-weighting unreliable information.
Let $\mathcal{C}_{\mathrm{high}} \subseteq \mathcal{C}$ denote the subset corresponding to high reliability, and define
\begin{equation}
\small
\mathcal{I}_{\mathrm{high}} = \{ i \mid c_i \in \mathcal{C}_{\mathrm{high}} \}.
\end{equation}
The final answer is then given by
\begin{equation} 
\small
\tilde{y} = f_\theta\big(x, \{r_i\}_{i \in \mathcal{I}_{\mathrm{high}}}\big),
\end{equation}
i.e., a selective projection onto the reliable steps in the reasoning trajectory. 
The Exploration-Commitment Decoupling mechanism enables the model to \emph{explore with awareness} during reasoning while \emph{commit cautiously} in the final answer, providing a principled approach to integrating calibration into generation without relying on post-hoc methods.
As a result, the model can retain useful partial knowledge without propagating potentially incorrect information.

\subsection{Calibration-Aware Generation via Structured Supervision}
\label{sec:methodology_supervision}

We instantiate the proposed Exploration-Commitment Decoupling paradigm through a structured supervision framework that jointly models reasoning, reliability estimation, and answer reconstruction. 
Our objective is to enable the model to simultaneously acquire (i) calibrated exploration, i.e., generating calibrated reasoning trajectories with reliability estimates, and (ii) selective commitment, i.e., forming answers grounded in reliable reasoning.

\paragraph{Reliability Calibration}

Given a reasoning model, we first generate original reasoning trajectories $r$ along with the final answers $y$ for each query $x$.
To derive supervision signals for reliability estimation, we assign a factuality score to each reasoning step $r_i$ using VeriScore \citep{song-etal-2024-veriscore}.
Specifically, each step $r_i$ is decomposed into a set of atomic claims $(a_{i,1}, \dots, a_{i,|\mathcal{A}_i|})$, which are verified against external evidence retrieved via a search engine.
The factuality score of $r_i$ is defined as the proportion of claims supported by the retrieved evidence:
\begin{equation}
\small
s_i = \frac{1}{|\mathcal{A}_i|} \sum_{j=1}^{|\mathcal{A}_i|} \mathbb{I}\big(a_{i,j} \text{ is supported}\big).
\end{equation}
We then discretize these scores into reliability labels $c_i \in \mathcal{C}$ via a bucketing function:
\begin{equation}
\small
\label{equ:bucketing_general}
c_i = \mathrm{Bucket}(s_i),
\end{equation}
where $\mathcal{C}$ denotes a predefined set of verbalized reliability levels. 
For instance, under a binary scheme:
\begin{equation}
\small
\label{equ:bucketing_binary}
\mathrm{Bucket}(s_i) =
\begin{cases}
\texttt{<reliable>}, & \text{if } s_i \ge \tau, \\
\texttt{<unreliable>}, & \text{if } s_i < \tau.
\end{cases}
\end{equation}
The motivation for discretization is two-fold.
(i) From a decision-theoretic perspective, it induces a thresholding rule over factuality scores, corresponding to an approximately Bayes-optimal decision rule under asymmetric utility (Appendix \ref{sec:appendix_bayes}).
(ii) From a representation perspective, expressing reliability as semantic tokens rather than continuous scalars better aligns with the language modeling objective \citep{DBLP:conf/icml/ChuangSGBB0Z25}, facilitating more effective learning of factuality awareness.

\paragraph{Answer Projection via Reliable Reasoning}

From augmented reasoning trajectories $(r, c)=\big((r_1, c_1), \dots, (r_{|\mathcal{R}|}, c_{|\mathcal{R}|})\big)$, we construct a refined target answer $\tilde{y}$ to facilitate selective commitment.
To this end, we prompt GPT-5 \citep{singh2025openaigpt5card} to revise the original answer $y$ conditioned on $(r, c)$, 
ensuring that the resulting answer depends only on reliable reasoning steps.

Specifically, the model is instructed to generate $\tilde{y}$ that preserves the structure and style of the original answer $y$, 
while enforcing selective commitment based on step-level reliability.
Content derived from unreliable steps (i.e., $\{r_i \mid c_i \in \mathcal{C}_{\mathrm{low}}\}$) is either suppressed or attenuated, whereas information supported by trustworthy steps is retained.
Importantly, the refinement process does not introduce any information that is not already present in $y$, ensuring that $\tilde{y}$ is obtained solely through selective filtering and reweighting of the original content.
Formally, $\tilde{y}$ can be interpreted as a projection of $y$ onto the subspace induced by $\{r_i\}_{i \in \mathcal{I}_{\mathrm{high}}}$, preserving linguistic fluency and coherence while constraining epistemic commitment.
This procedure yields training tuples $\mathcal{D}_{\mathrm{CASS}}$ of the form $(x, (r, c), \tilde{y})$, which encode both calibrated exploration and selective commitment under a unified supervision signal.

\paragraph{Projection Quality}

We conduct a quality check to assess whether the proposed projection mechanism achieves its intended goals: (i) improving factuality by filtering unreliable content, and (ii) preserving faithfulness without introducing new knowledge.
We sample 1,000 instances and compare answers before and after projection.
The average factuality score, measured by VeriScore, improves from 67.79\% to 79.80\%, indicating effective removal of information derived from unreliable reasoning.
We further examine whether projection introduces new information beyond the reasoning $r$ and original answer $y$. Using GPT-5 as a verifier, we find that 96\% of projected answers are fully supported by the original reasoning and answer, suggesting that projection largely preserves fidelity and exhibits minimal knowledge expansion.

\paragraph{Training Objective}

We train the model to jointly generate reliability-aware reasoning and the refined final answer under the proposed \textbf{Calibration-Aware Structured Supervision (CASS)} scheme:
\begin{equation}
\small
\mathcal{L}_{\mathrm{CASS}} =
\mathbb{E}_{(x, r, c, \tilde{y}) \sim \mathcal{D}_{\mathrm{CASS}}}
\left[
-\log p_\theta(r, c \mid x)
- \log p_\theta(\tilde{y} \mid x, r, c)
\right],
\end{equation}
which aligns generation with the proposed paradigm:
(i) the first term encourages calibrated reasoning with explicit reliability estimates (i.e., \emph{knowing what the model knows}), and
(ii) the second term promotes answer construction grounded in reliable reasoning (i.e., \emph{acting on what the model knows}).

\subsection{Refining Calibration-Aware Generation via On-Policy Distillation}
\label{sec:methodology_distillation}

While Calibration-Aware Structured Supervision enables the model to acquire an initial calibration-aware generation policy, smaller models often struggle with reliability estimation and robust selective commitment. To this end, we employ on-policy distillation \citep{DBLP:conf/iclr/AgarwalVZSGGB24,lu2025onpolicydistillation} to transfer calibration-aware generation behaviors from a stronger teacher model to the student.

\paragraph{On-Policy Student Rollout}

Given an input query $x$, the student model $\theta_s$ generates a reliability-aware reasoning trajectory along with a committed answer:
\begin{equation}
\small
z^s = (r^s, c^s, y^s) \sim p_{\theta_s}(\cdot \mid x).
\end{equation}
The on-policy rollout exposes the student to its own generation distribution, capturing realistic failure modes such as miscalibrated reliability and suboptimal commitment decisions during inference.

\paragraph{Calibration-Aware Policy Distillation}

To refine the student policy, we leverage a stronger teacher model $\theta_t$ to provide token-level guidance conditioned on the student's own generation. 
Specifically, we perform \textbf{Calibration-Aware Policy Distillation (CAPD)} by minimizing the divergence between the student and teacher predictive distributions:
\begin{equation}
\small
\mathcal{L}_{\mathrm{CAPD}}
=
\mathbb{E}_{x \sim \mathcal{D}}
\;
\mathbb{E}_{z^s \sim p_{\theta_s}(\cdot \mid x)}
\sum_{t=1}^{|z^s|}
D_{\mathrm{KL}}\Big(
p_{\theta_s}(\cdot \mid x, z^s_{<t})
\;\|\;
p_{\theta_t}(\cdot \mid x, z^s_{<t})
\Big),
\end{equation}
which improves calibration-aware generation by (i) correcting miscalibrated reliability estimates, and (ii) enabling more robust selective commitment under the student’s own generation distribution.

\section{Experiments}

\subsection{Experimental Setup}

\paragraph{Models and Datasets}

Our experiments cover 5 LLMs with varying scales and architectures: Llama-3.2-3B \citep{grattafiori2024llama3herdmodels}, Llama-3.1-8B, Qwen3-4B \citep{yang2025qwen3technicalreport}, Qwen3-8B, and Qwen3-14B. 
As the Llama series lacks native reasoning abilities, we adopt a cold-start initialization to elicit such behavior.
We evaluate our methods on 5 widely used long-form factuality benchmarks: AlpacaFact \citep{DBLP:conf/nips/DuboisLTZGBGLH23}, Biography \citep{DBLP:conf/emnlp/MinKLLYKIZH23}, FactBench \citep{fatahi-bayat-etal-2025-factbench}, Factory \citep{chen2025factorychallenginghumanverifiedprompt}, and FAVA \citep{mishra2024finegrained}. 
Additional details are in Appendix \ref{sec:appendix_implementation}.

\paragraph{Baselines and Metrics}

We compare our approach against two categories of baselines: 
(i) abstention-based methods \citep{xue-etal-2025-ualign,zhang-etal-2024-r}, and 
(ii) factuality-optimized methods \citep{chen2025learningreasonfactuality,chen2025traintruthskillsbinary}. 
For the latter, we consider two representative approaches: RL with a continuous VeriScore reward \citep{chen2025learningreasonfactuality}, and RL with a Binary Retrieval-Augmented Reward (Binary RAR) \citep{chen2025traintruthskillsbinary}.
We adopt VeriScore \citep{song-etal-2024-veriscore} as the primary metric for evaluating factuality and helpfulness in long-form responses, which follows a three-stage pipeline: (i) claim extraction, (ii) evidence retrieval, and (iii) claim verification. 
Given a response $y$, a set of claims $\mathcal{A} = \{a_1, \dots, a_{|\mathcal{A}|}\}$ is first extracted. 
Each claim is then associated with retrieved evidence and subsequently verified for factual support.
Factuality is defined as $P(y) = S(y) / |\mathcal{A}|$, where $S(y)$ denotes the number of supported claims. 
Helpfulness is defined as $R(y) = \min(S(y) / K, 1)$, where $K$ represents the expected number of correct claims for the target domain. 
The final VeriScore is $F_1 = 2 P(y) R(y) / (P(y) + R(y))$ if $S(y) > 0$, and $0$ otherwise.
More details are in Appendix \ref{sec:appendix_implementation}.

\subsection{Implementation Details}

We utilize GPT-5 to filter a diverse, high-quality set of prompts reflecting realistic user interactions from ELI5 \citep{fan-etal-2019-eli5}, yielding 5K prompts for Calibration-Aware Structured Supervision and 3K for Calibration-Aware Policy Distillation.
Experiments are conducted on 8 H100 GPUs. For the Llama series, we adopt a cold-start training with a learning rate of $2 \times 10^{-5}$, a per-GPU batch size of 4, and train for 1 epoch.
For Calibration-Aware Structured Supervision (\S\ref{sec:methodology_supervision}), we select the optimal threshold $\tau^*$ for bucketing via grid search, with detailed ablations on the choice of $\tau$ provided in Appendix \ref{sec:appendix_threshold_ablation}.
Models are trained for 2 epochs with a per-GPU batch size of 4. The learning rates are set to $2 \times 10^{-5}$ for Llama-3.2-3B, and $1 \times 10^{-5}$ for Llama-3.1-8B, Qwen3-4B, Qwen3-8B, and Qwen3-14B.
For Calibration-Aware Policy Distillation (\S\ref{sec:methodology_distillation}), the largest models in each series (i.e., Llama-3.1-8B and Qwen3-14B) are used as teachers to distill smaller students. 
The training is performed for 2 epochs with a per-GPU batch size of 2. The learning rates are set to $8 \times 10^{-6}$ for Llama-3.2-3B, and $6 \times 10^{-6}$ for Qwen3-4B, and Qwen3-8B.
For evaluation, GPT-5 is used as the backbone model for VeriScore.
Further implementation details are deferred to Appendix \ref{sec:appendix_implementation}.

\subsection{Main Results}

\begin{table}
\centering
\small
\caption{Main results on five long-form factuality benchmarks. CASS significantly improves performance over baselines across all models, and further gains are achieved with CAPD, demonstrating the effectiveness of calibration-aware generation. Full results in Appendix \ref{sec:appendix_main_results}.}
\label{tab:main_results}
\adjustbox{width=0.86\textwidth}{
\begin{tabular}{l*{3}{S[table-format=2.2, table-column-width=1.5cm]}*{3}{S[table-format=2.2, table-column-width=1.25cm]}}
\toprule
{\textbf{Models}} & {\textbf{AlpacaFact}} & {\textbf{Biography}} & {\textbf{FactBench}} & {\textbf{Factory}} & {\textbf{FAVA}} & {\textbf{AVG}} \\

\midrule

\textbf{Qwen3-4B} & 67.93 & 35.27 & 71.25 & 64.37 & 64.18 & 60.60 \\
+ Abstention & 46.00 & 23.40 & 45.92 & 25.91 & 48.13 & 37.87 \\
+ RL (VeriScore) & 68.29 & 32.00 & 72.35 & 61.58 & 65.45 & 59.93 \\
+ RL (Binary RAR) & 67.94 & 32.43 & 71.51 & 62.35 & 64.43 & 59.73 \\
\rowcolor{mygray} + CASS & 71.16 & 39.05 & 74.02 & 65.41 & 66.52 & 63.23 \\
\rowcolor{mygray} + CASS + CAPD & \textbf{76.44} & \textbf{40.62} & \textbf{75.38} & \textbf{67.30} & \textbf{70.76} & \textbf{66.10} \\

\midrule
\addlinespace

\textbf{Qwen3-8B} & 73.65 & 39.90 & 75.18 & 68.19 & 71.21 & 65.63 \\
+ Abstention & 33.70 & 21.93 & 28.37 & 22.68 & 44.50 & 30.24 \\
+ RL (VeriScore) & 74.09 & 40.04 & 73.80 & 65.72 & 72.13 & 65.16 \\
+ RL (Binary RAR) & 73.79 & 40.05 & 75.22 & 67.70 & 72.04 & 65.76 \\
\rowcolor{mygray} + CASS & 75.47 & \textbf{49.73} & 77.61 & 69.34 & 73.42 & 69.11 \\
\rowcolor{mygray} + CASS + CAPD & \textbf{78.24} & 49.64 & \textbf{79.79} & \textbf{70.37} & \textbf{75.13} & \textbf{70.63} \\

\midrule
\addlinespace

\textbf{Qwen3-14B} & 76.05 & 43.47 & 78.26 & 69.02 & 70.67 & 67.49 \\
+ Abstention & 55.61 & 32.23 & 52.58 & 21.42 & 59.23 & 44.21 \\
+ RL (VeriScore) & 77.82 & 47.66 & 79.68 & 67.25 & 73.35 & 69.15 \\
+ RL (Binary RAR) & 76.33 & 46.50 & 76.95 & 68.49 & 74.11 & 68.48 \\
\rowcolor{mygray} + CASS & \textbf{80.77} & \textbf{54.70} & \textbf{81.87} & \textbf{71.29} & \textbf{76.63} & \textbf{73.05} \\

\midrule
\addlinespace

\textbf{Llama-3.1-8B} & 71.55 & 48.18 & 69.03 & 54.22 & 64.59 & 61.51 \\
+ Abstention & 22.51 & 33.94 & 16.04 & 20.13 & 28.24 & 24.17 \\
+ RL (VeriScore) & 75.60 & 54.99 & 73.30 & 60.95 & 74.89 & 67.95 \\
+ RL (Binary RAR) & 76.16 & 55.66 & 73.52 & 61.01 & 74.15 & 68.10 \\
\rowcolor{mygray} + CASS & \textbf{77.05} & \textbf{61.10} & \textbf{77.52} & \textbf{66.24} & \textbf{77.03} & \textbf{71.79} \\

\bottomrule
\end{tabular}
}
\end{table}

Table \ref{tab:main_results} summarizes the results on five long-form factuality benchmarks evaluated by VeriScore. Several key observations can be drawn.
\textbf{(i) Abstention performs poorly in long-form generation}, as directly refusing to answer substantially harms helpfulness, leading to large drops in overall scores.
\textbf{(ii) CASS, our instantiation of CAG, consistently outperforms all baselines}, with notable improvements such as boosting Llama-3.1-8B from 61.51 to 71.79 (+10.3). Similar gains are observed across models, demonstrating strong generalization. Importantly, \textbf{CASS does not introduce new knowledge or external information}, indicating that the improvements arise purely from better calibration and generation policies.
\textbf{(iii) CAPD provides additional gains for smaller models}, further validating its effectiveness in refining calibration-aware generation.
Full results for all models, along with fine-grained analyses of factuality and helpfulness, are provided in Appendix \ref{sec:appendix_main_results}.

\section{Analysis}

To better understand the effectiveness of CAG, we conduct analyses along several key dimensions.
We first perform an ablation study (\S\ref{sec:analysis_ablation}) to evaluate the roles of calibrated exploration and selective commitment.
We then assess its empirical performance from three perspectives: (i) generalization across diverse generation settings (\S\ref{sec:analysis_generalization}), (ii) compatibility with Retrieval-Augmented Generation (RAG) (\S\ref{sec:analysis_rag}), and (iii) efficiency in decoding time and token usage (\S\ref{sec:analysis_efficiency}).
Building on these observations, we further investigate the underlying mechanisms of CAG by analyzing (iv) calibration quality (\S\ref{sec:analysis_calibration}) and (v) how calibrated reasoning shapes answer organization (\S\ref{sec:analysis_answer_organization}).

\subsection{Ablation Study}
\label{sec:analysis_ablation}

We conduct ablation experiments to assess the contributions of calibrated exploration and selective commitment. Specifically, we consider two variants: (i) w/o Calibrated Exploration, which removes reliability estimates during reasoning, and (ii) w/o Selective Commitment, which adheres to the original final answer without leveraging reliability signals to filter or prioritize reasoning.
As shown in Table \ref{tab:analysis_ablation}, removing either component leads to consistent performance degradation across all benchmarks and model scales, indicating that both calibrated exploration and selective commitment are crucial for effective calibration-aware generation.

\begin{table}[!tbp]
\centering
\small
\caption{Ablation study. Removing reliability calibration or selective commitment consistently degrades performance, underscoring their importance in calibration-aware generation.}
\label{tab:analysis_ablation}
\adjustbox{width=0.84\textwidth}{
\begin{tabular}{l*{3}{S[table-format=2.2, table-column-width=1.6cm]}*{2}{S[table-format=2.2, table-column-width=1.4cm]}}
\toprule
{\textbf{Models}} & {\textbf{AlpacaFact}} & {\textbf{Biography}} & {\textbf{FactBench}} & {\textbf{Factory}} & {\textbf{FAVA}} \\

\midrule

w/o Calibrated Exploration & 77.75 & 48.12 & 79.46 & 70.25 & 73.14 \\
w/o Selective Commitment & 77.03 & 46.47 & 76.08 & 68.68 & 71.69 \\
\rowcolor{mygray} \textbf{Qwen3-14B + CASS} & \textbf{80.77} & \textbf{54.70} & \textbf{81.87} & \textbf{71.29} & \textbf{76.63} \\

\midrule

w/o Calibrated Exploration & 75.73 & 55.70 & 71.60 & 62.56 & 73.38 \\
w/o Selective Commitment & 74.01 & 50.94 & 71.90 & 59.88 & 68.20 \\
\rowcolor{mygray} \textbf{Llama-3.1-8B + CASS} & \textbf{77.05} & \textbf{61.10} & \textbf{77.52} & \textbf{66.24} & \textbf{77.03} \\

\bottomrule
\end{tabular}
}
\end{table}

\subsection{Generalization to Diverse Generation Scenarios}
\label{sec:analysis_generalization}

\begin{figure}[!htbp]
\centering
\adjustbox{width=0.9\textwidth}{
\begin{minipage}{0.32\textwidth}
    \centering
    \includegraphics[width=\linewidth]{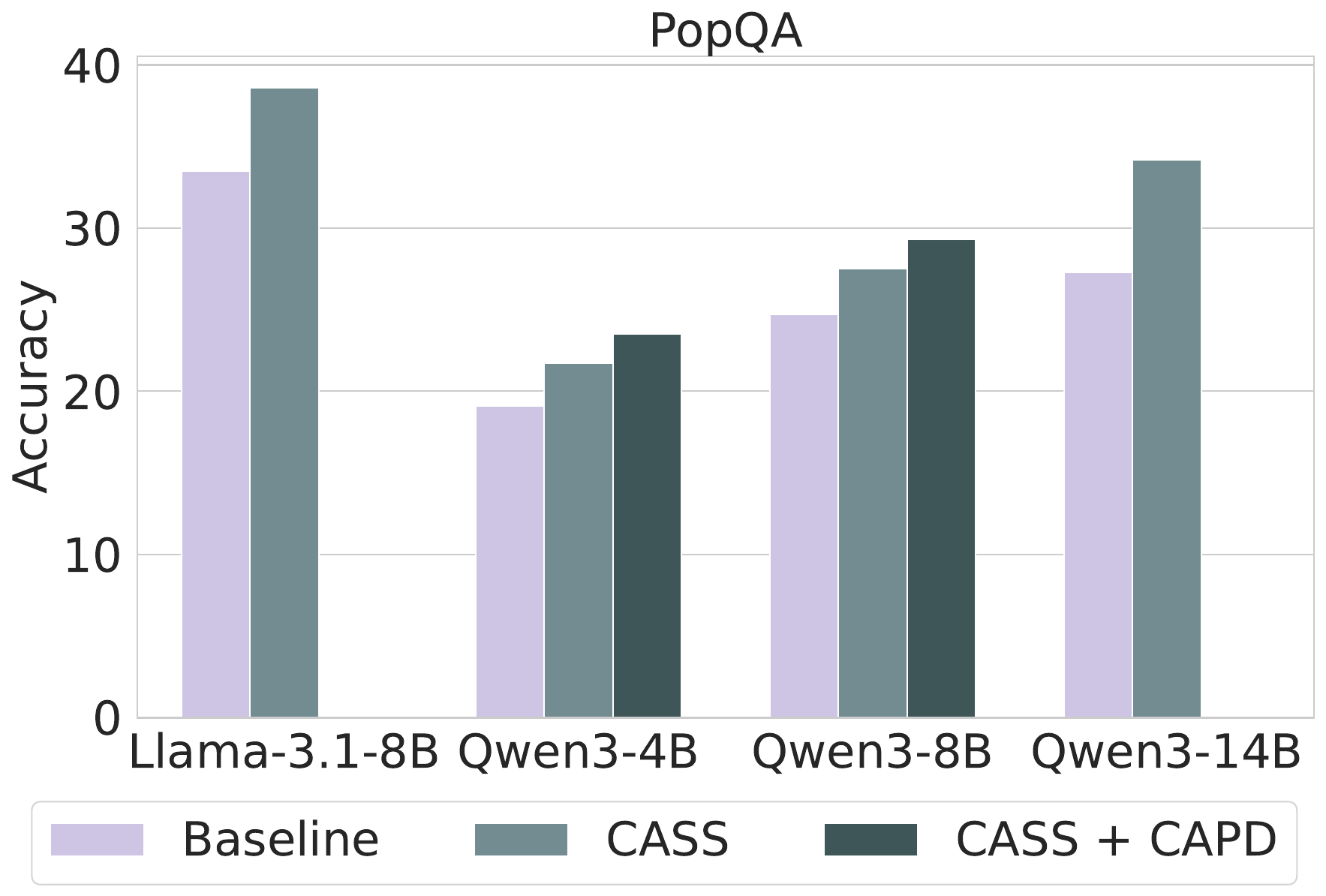}
\end{minipage}
\hfill
\begin{minipage}{0.32\textwidth}
    \centering
    \includegraphics[width=\linewidth]{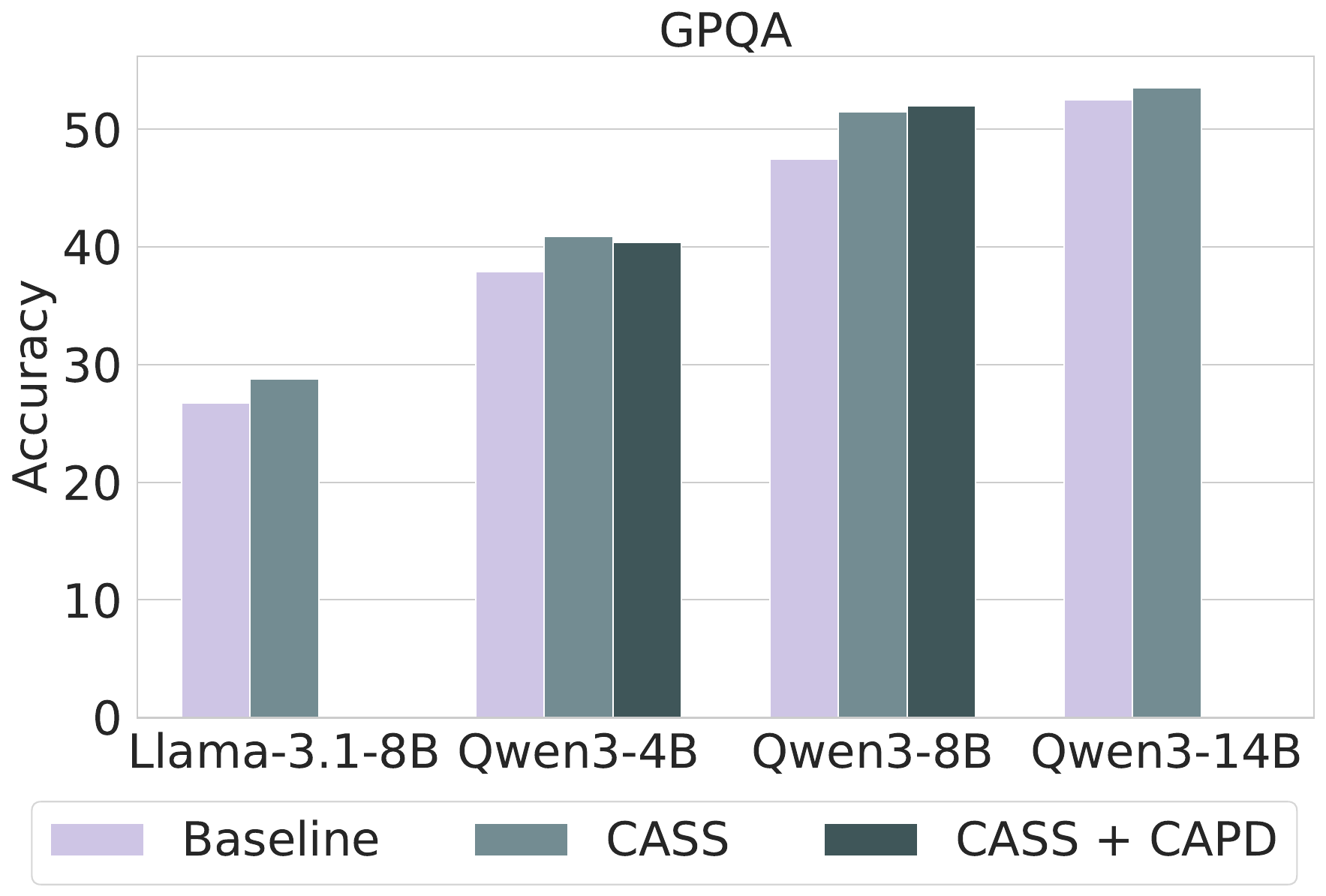}
\end{minipage}
\hfill
\begin{minipage}{0.32\textwidth}
    \centering
    \includegraphics[width=\linewidth]{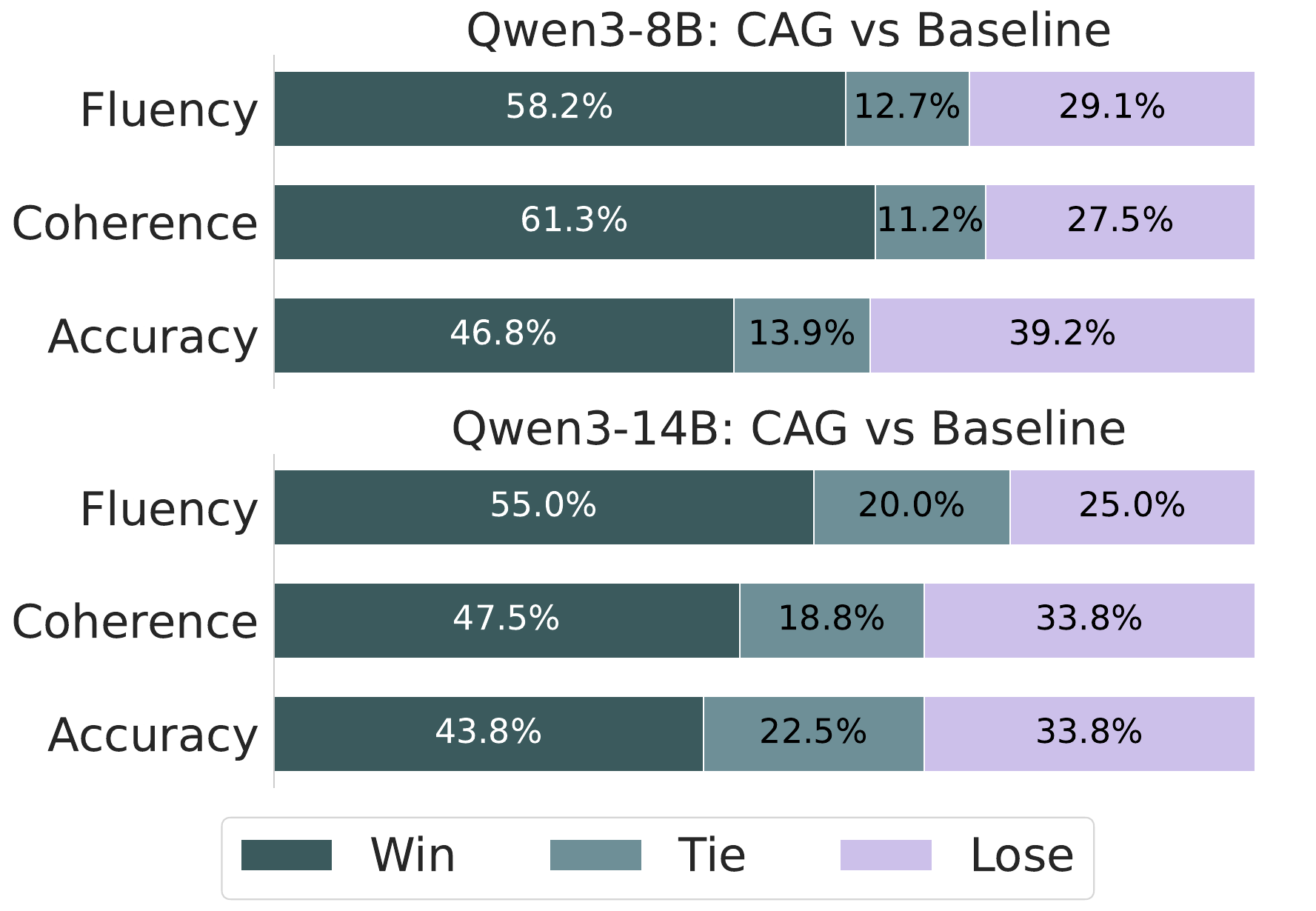}
\end{minipage}
}
\caption{Performance comparison across PopQA, GPQA, and Vicuna QA benchmarks. CASS and CAPD consistently outperform baselines, indicating strong general capabilities.}
\label{fig:analysis_generalization}
\end{figure}

In this section, we evaluate the general capabilities of CAG across two representative scenarios: knowledge-intensive question answering and open-ended chatbot settings.
(i) For knowledge-intensive QA, experiments are conducted on PopQA \citep{DBLP:conf/acl/MallenAZDKH23} and GPQA \citep{rein2024gpqa}. As shown in Figure \ref{fig:analysis_generalization}, both CASS and CAPD consistently outperform baseline methods across the datasets, suggesting that calibrated reasoning enables more effective knowledge recall, organization, and application.
(ii) We further examine CAG in an open-ended chatbot setting using the Vicuna QA benchmark \citep{DBLP:conf/nips/ZhengC00WZL0LXZ23}, with evaluation criteria focusing on fluency, coherence, and accuracy. The results show that CASS and CAPD achieve superior performance in the majority of cases across all three dimensions.
Overall, the findings demonstrate that CAG generalizes robustly across diverse scenarios, yielding consistent improvements in response quality.

\subsection{Compatibility with Retrieval-Augmented Generation}
\label{sec:analysis_rag}

\begin{table}[!tbp]
\centering
\small
\caption{Performance comparison with and without RAG. CASS and CAPD consistently improve performance, highlighting effectiveness in both standalone and retrieval-augmented settings.}
\label{tab:analysis_rag}
\adjustbox{width=0.85\textwidth}{
\begin{tabular}{l*{6}{S[table-format=2.2, table-column-width=1.4cm]}}
\toprule
\multirow{2}{*}{\textbf{Models}}  & \multicolumn{2}{c}{\textbf{Biography}} & \multicolumn{2}{c}{\textbf{FAVA}} & \multicolumn{2}{c}{\textbf{AVG}}\\
\cmidrule(lr){2-3} \cmidrule(lr){4-5} \cmidrule(lr){6-7}
 & {w/o RAG} & {w/ RAG} & {w/o RAG} & {w/ RAG} & {w/o RAG} & {w/ RAG} \\
\midrule

\textbf{Qwen3-8B} & 39.90 & 84.40 & 71.21 & 73.16 & 55.55 & 78.78 \\
\rowcolor{mygray} + CASS & \textbf{49.73} & 86.45 & 73.42 & 76.39 & 61.58 & 81.42 \\
\rowcolor{mygray} + CASS + CAPD & 49.64 & \textbf{88.22} & \textbf{75.13} & \textbf{81.37} & \textbf{62.38} & \textbf{84.80} \\

\midrule

\textbf{Qwen3-14B} & 43.47 & 87.62 & 70.67 & 81.84 & 57.07 & 84.73 \\
\rowcolor{mygray} + CASS & \textbf{54.70} & \textbf{89.87} & \textbf{76.63} & \textbf{84.96} & \textbf{65.66} & \textbf{87.41} \\

\midrule

\textbf{Llama-3.1-8B} & 48.18 & 81.31 & 64.59 & 69.01 & 56.39 & 75.16 \\
\rowcolor{mygray} + CASS & \textbf{61.10} & \textbf{88.30} & \textbf{77.03} & \textbf{81.23} & \textbf{69.06} & \textbf{84.77} \\

\bottomrule
\end{tabular}
}
\end{table}

We evaluate the compatibility of our method with RAG \citep{wang2025chainofretrievalaugmentedgeneration}, where models are provided with external evidence during inference. As shown in Table \ref{tab:analysis_rag}, CASS and CAPD consistently improve performance over baselines in both RAG and non-RAG settings across all model families and benchmarks. The results suggest that the benefits of calibration-aware generation are not dependent on the availability of external knowledge, but instead arise from improved calibrated reasoning and selective commitment. Moreover, the gains persist even in the presence of strong retrieval signals, indicating that CAG enables models to more effectively leverage retrieved evidence.

\subsection{Efficiency}
\label{sec:analysis_efficiency}

We evaluate the efficiency of CAG in terms of decoding time and token usage.
We report the average decoding time per sample (in seconds) and the relative change in generated token length, which are further decomposed into \textit{thinking} and \textit{answer} components.
As shown in Table \ref{tab:efficiency}, CAG consistently reduces decoding time across all models, achieving a reduction of up to 37.36\%. 
While the thinking length remains largely stable, the answer segment is significantly shortened, leading to a notable reduction in overall token usage. 
By maintaining sufficient deliberation while reducing unnecessary or incorrect outputs, CAG not only enhances factual accuracy but also improves generation efficiency.

\begin{table}[!tbp]
\centering
\small
\caption{Efficiency comparison in terms of decoding time and token usage. The average decoding time per sample (in seconds) and the relative change in generated token length are reported.}
\label{tab:efficiency}
\adjustbox{width=0.84\textwidth}{
\begin{tabular}{l*{6}{S[table-format=2.2, table-column-width=1.5cm]}}
\toprule
\multirow{2}{*}{\textbf{Models}} & \multicolumn{3}{c}{\textbf{Decoding Time (s)}} & \multicolumn{3}{c}{\textbf{$\Delta$Token Length (\%)}} \\
\cmidrule(lr){2-4} \cmidrule(lr){5-7}
& {\textbf{Baseline}} & {\textbf{CAG}} & {\textbf{$\Delta$ (\%)}} 
& {\textbf{Thinking}} & {\textbf{Answer}} & {\textbf{Total}} \\
\midrule

\textbf{Qwen3-4B}  & 62.91 & 47.03 & -25.24\% & -12.83\% & -56.07\% & -31.43\% \\
\textbf{Qwen3-8B}  & 58.46 & 36.63 & -37.36\% & -11.74\% & -63.41\% & -35.00\% \\
\textbf{Qwen3-14B} & 50.57 & 40.11 & -20.67\% & -3.05\%  & -60.86\% & -29.86\% \\

\bottomrule
\end{tabular}
}
\end{table}

\subsection{How Well-Calibrated is the Model?}
\label{sec:analysis_calibration}

\begin{figure}
\centering
\adjustbox{width=0.8\textwidth}{
\begin{minipage}{0.45\textwidth}
    \centering
    \includegraphics[width=\linewidth]{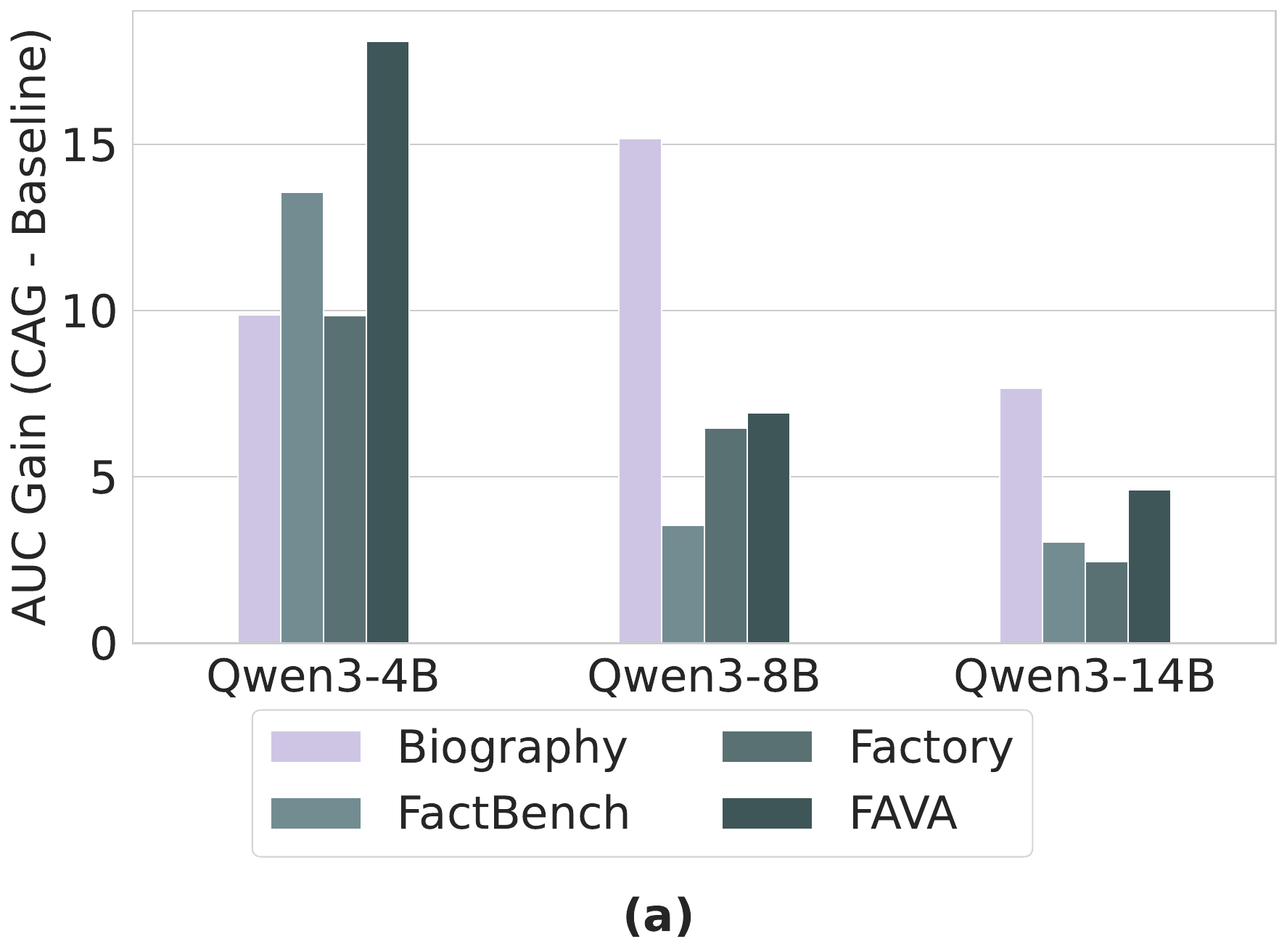}
\end{minipage}
\hfill
\begin{minipage}{0.52\textwidth}
    \centering
    \includegraphics[width=\linewidth]{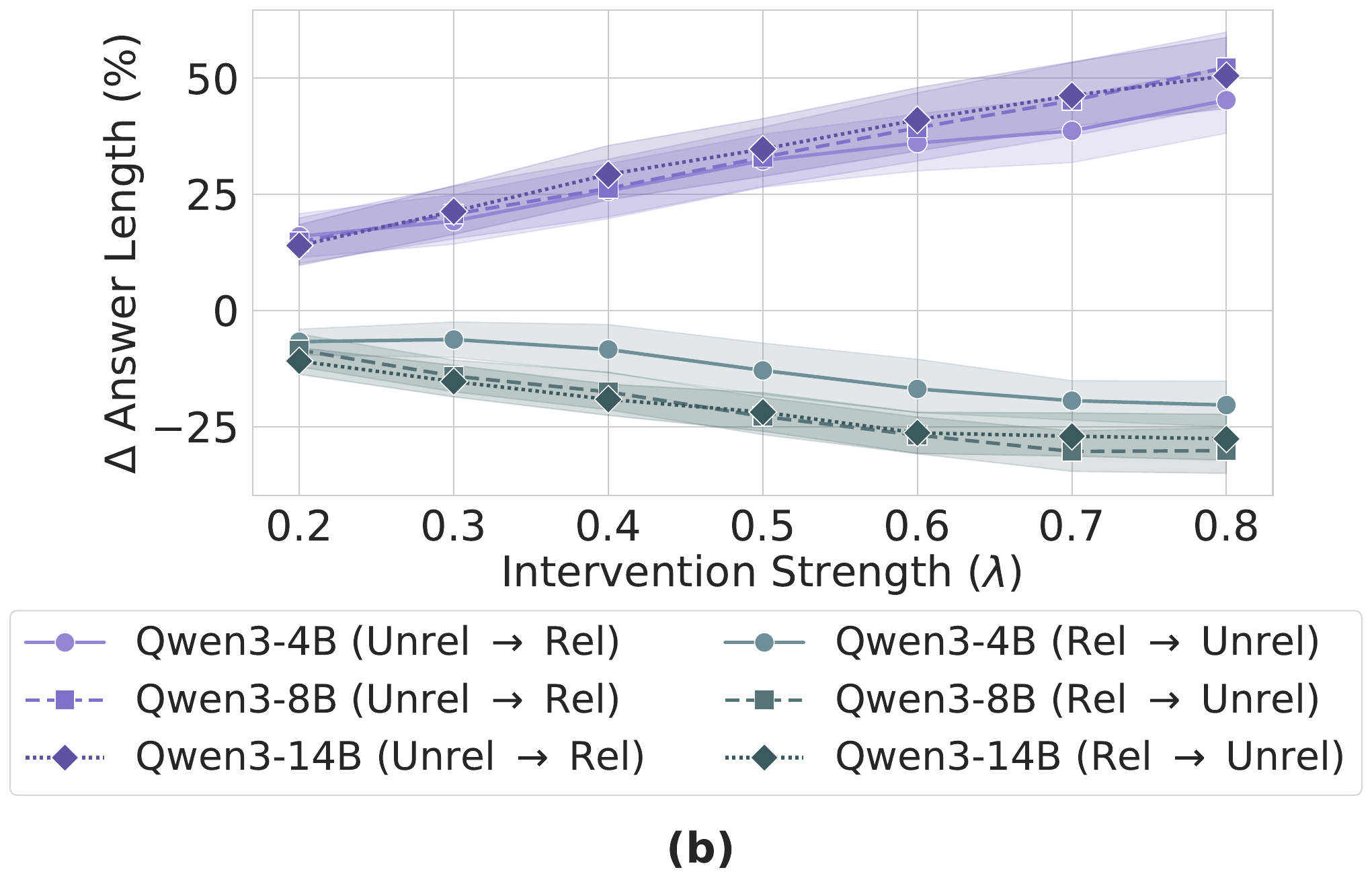}
\end{minipage}
}
\caption{Calibration performance and its effect on answer organization.
(a) Calibration quality measured by absolute AUC gain. CAG consistently outperforms the baseline, indicating better alignment between predicted reliability and factual correctness.
(b) Reliability intervention affects answer length: treating unreliable steps as reliable (\textbf{Unrel $\rightarrow$ Rel}) lengthens responses, while the reverse (\textbf{Rel $\rightarrow$ Unrel}) shortens them, showing calibration regulates information bandwidth. Shaded regions indicate 95\% confidence intervals.}
\label{fig:analysis_calibration_intervention}
\end{figure}

To assess the calibration quality of CAG, we analyze the alignment between predicted reliability and the factual correctness of intermediate reasoning steps.
Experiments are conducted on the Biography and FAVA benchmarks, with calibration performance quantified using AUC, which measures the correlation between estimated reliability and ground-truth factuality scores. 
As a baseline, we consider the vanilla model of the same size under an LLM-as-a-judge setting without calibration-aware mechanisms.
As shown in Figure \ref{fig:analysis_calibration_intervention}(a), CAG consistently achieves substantial improvements in AUC over the baseline, indicating a significantly stronger alignment between reliability estimates and actual correctness.
The results suggest that \textbf{CAG enhances the model's awareness of its own knowledge}.
Improved calibration allows for more accurate discrimination between reasoning steps, leading to more effective use of evidence and more reliable outputs.

\subsection{How Calibration Shapes Answer Organization}
\label{sec:analysis_answer_organization}

In this section, we investigate how calibrated reasoning shapes answer organization. Our key observation is that \textbf{calibrated reliability regulates the effective information bandwidth of the final output}.
We perform controlled interventions in which the reliability of each reasoning step is independently flipped with probability $\lambda$. 
Two settings are considered: \textbf{(i) Unrel $\rightarrow$ Rel}, where unreliable steps are treated as reliable, and \textbf{(ii) Rel $\rightarrow$ Unrel}, where reliable steps are instead marked as unreliable. 
We then measure the resulting changes in answer length.
As shown in Figure \ref{fig:analysis_calibration_intervention}(b), increasing $\lambda$ in setting (i) leads to a substantial increase in answer length, whereas in setting (ii) it results in a decrease, suggesting that when more reasoning steps are (incorrectly) considered reliable, the model is more likely to include additional content. Conversely, lowering perceived reliability suppresses extended generation.
Overall, calibration plays a causal role in answer organization. 
By governing which reasoning steps are trusted, it shapes both \emph{what the model generates} and \emph{how much it generates}, thereby promoting more reliable responses.

\section{Related Work}

Despite their impressive capabilities across diverse tasks, Large Reasoning Models \citep{singh2025openaigpt5card} are prone to hallucination—outputs that appear plausible yet are factually incorrect. 
Prior work on improving factuality can be broadly categorized into two main directions: (i) abstention-based methods and (ii) factuality-optimized methods.
Abstention-based approaches \citep{wu2026mitigatingllmhallucinationbehaviorally,an2025teachingllmsabstainfinegrained,DBLP:conf/acl/YuanJ00XLHT25} encourage models to abstain when they are likely to be incorrect \citep{ren2025knowrlexploringknowledgeablereinforcement,xue-etal-2025-ualign,luo2026pathwaystruthfulnessintrinsicencoding}. 
Factuality-optimized methods improve factuality by directly optimizing model behavior using specialized training objectives, including supervised fine-tuning \citep{DBLP:conf/acl/XieZPJMFTWFRFZ25,DBLP:conf/nips/LinGOXLY024} and reinforcement learning with factuality-oriented rewards \citep{chen2025learningreasonfactuality,chen2025traintruthskillsbinary,wu2026mitigatingllmhallucinationbehaviorally}.
Additionally, a range of post-hoc methods have been proposed, such as retrieval-augmented generation \citep{DBLP:conf/iclr/AsaiWWSH24}, prompting techniques \citep{luo2024halludiallargescalebenchmarkautomatic}, and decoding algorithms \citep{DBLP:conf/emnlp/ChuangQHKKG24,DBLP:conf/acl/000100PWW25}.
Despite their differences, these approaches adhere to a coupled exploration–commitment paradigm, where intermediate reasoning is unconditionally carried into the final output, preventing fine-grained control over information selection and integration and consequently limiting their effectiveness in long-form factuality.
Compared to prior approaches, our method explicitly decouples exploration from commitment. Rather than globally suppressing or optimizing outputs, it employs calibration-aware generation to enable fine-grained selection of trustworthy information, thereby improving factuality without sacrificing utility.

\section{Conclusion}

In this paper, we propose an \textbf{Exploration–Commitment Decoupling} paradigm to improve long-form factuality by disentangling knowledge exploration from final answer commitment. 
We instantiate the paradigm with \textbf{Calibration-Aware Generation (CAG)}, a framework that integrates calibrated exploration and selective commitment, enabling models to perform end-to-end, calibration-aware generation by estimating step-level reliability and selectively incorporating trustworthy reasoning into final outputs.
Experiments demonstrate consistent factuality improvements across models and settings, offering directions for more trustworthy and self-aware generative systems.



\clearpage

\bibliographystyle{unsrtnat}
\bibliography{references}

\clearpage
\appendix

\section{Approximate Bayes-Optimal Selection via Score Thresholding}
\label{sec:appendix_bayes}

In this section, we provide a decision-theoretic justification for reliability bucketing. 
We show that (i) thresholding posterior correctness probabilities is Bayes-optimal under asymmetric utility, and (ii) thresholding an accurate estimate of these probabilities yields a near-optimal decision rule.

\paragraph{Setup}

Consider a reasoning trace consisting of $|\mathcal{R}|$ steps $\{r_i\}_{i=1}^{|\mathcal{R}|}$.
For each step $r_i$, let $z_i \in \{0,1\}$ denote its latent factual correctness, where $z_i = 1$ indicates that the step is correct.

Let
\begin{equation}
p_i := \mathbb{P}(z_i = 1 \mid r_i)
\end{equation}
be the posterior probability that step $r_i$ is correct, and let $d_i \in \{0,1\}$ denote whether the model commits to this step.

\paragraph{Utility}

We consider the asymmetric utility
\begin{equation}
U(d_i, z_i) =
\begin{cases}
+u_1, & d_i = 1,\, z_i = 1, \\
-u_2, & d_i = 1,\, z_i = 0, \\
0,    & d_i = 0,
\end{cases}
\end{equation}
where $u_1 > 0$ and $u_2 > 0$.

Define the conditional expected utility
\begin{equation}
\bar U(d_i \mid r_i) := \mathbb{E}[U(d_i, z_i)\mid r_i].
\end{equation}

\paragraph{Bayes-Optimal Decision Rule}

The conditional expected utility of committing is
\begin{equation}
\bar U(1 \mid r_i) = u_1 p_i - u_2(1-p_i),
\end{equation}
while discarding yields
\begin{equation}
\bar U(0 \mid r_i)=0.
\end{equation}

Therefore, a Bayes-optimal decision rule is
\begin{equation}
d_i^* = \mathbb{I}(p_i \ge \tau^*),
\qquad
\tau^* = \frac{u_2}{u_1 + u_2}.
\end{equation}

Thus, optimal selection reduces to thresholding the posterior correctness probability.

\paragraph{Approximate Optimality via Proxy-Based Decisions}

In practice, the true posterior $p_i$ is unknown. Instead, we observe a score $s_i \in [0,1]$ (e.g., from an external verifier) that estimates $p_i$.

Assume $s_i$ is uniformly accurate:
\begin{equation}
|s_i - p_i| \le \epsilon.
\end{equation}

We define the proxy-based decision rule
\begin{equation}
\hat d_i = \mathbb{I}(s_i \ge \tau^*).
\end{equation}

\paragraph{Regret Bound}

Define the conditional value gap between committing and discarding:
\begin{equation}
V_i := \bar U(1\mid r_i)-\bar U(0\mid r_i)
= u_1 p_i - u_2(1-p_i)
= (u_1+u_2)(p_i-\tau^*).
\end{equation}
Hence the optimal rule commits iff $V_i\ge 0$, equivalently iff $p_i\ge \tau^*$, while the proxy rule commits iff $s_i\ge \tau^*$.

Let the conditional regret be
\begin{equation}
\Delta_i := \bar U(d_i^*\mid r_i)-\bar U(\hat d_i\mid r_i).
\end{equation}

If $\hat{d}_i = d_i^*$, then $\Delta_i = 0$.  
Otherwise, $p_i$ and $s_i$ lie on opposite sides of $\tau^*$, which implies
\begin{equation}
|p_i-\tau^*| \le |p_i-s_i| \le \epsilon.
\end{equation}

Therefore,
\begin{equation}
\Delta_i 
= |V_i|
= (u_1 + u_2)|p_i - \tau^*|
\le (u_1 + u_2)\epsilon.
\end{equation}

Thus, thresholding an $\epsilon$-accurate estimate of the posterior yields an $\mathcal{O}(\epsilon)$-optimal decision rule.

\section{Implementation Details}
\label{sec:appendix_implementation}

\subsection{Models}
\label{sec:appendix_models}

Our experiments span 5 LLMs that differ in both scale and architectural design, including Llama-3.2-3B \citep{grattafiori2024llama3herdmodels}, Llama-3.1-8B, Qwen3-4B \citep{yang2025qwen3technicalreport}, Qwen3-8B, and Qwen3-14B. 

\subsection{Datasets}
\label{sec:appendix_datasets}

We evaluate CAG on five long-form factuality benchmarks spanning diverse domains and question types.

\textbf{AlpacaFact} comprises 253 fact-seeking instructions drawn from the AlpacaFarm dataset \citep{DBLP:conf/nips/DuboisLTZGBGLH23}, which contains 805 real-world user instructions in total. We follow the selection protocol of \citet{DBLP:conf/nips/LinGOXLY024}.

\textbf{Biography} \citep{DBLP:conf/emnlp/MinKLLYKIZH23} contains 183 questions about public figures sourced from Wikipedia, targeting the factuality of long-form generation in a knowledge-intensive setting.

\textbf{FactBench} \citep{fatahi-bayat-etal-2025-factbench} is derived from LMSYS-Chat-1M \citep{DBLP:conf/iclr/ZhengC0LZW00LXG24} by filtering LLM responses based on hallucination scores, resulting in prompts that are particularly challenging for factual correctness. We adopt its most difficult tier, FactBench-Hard, which comprises 532 questions.

\textbf{Factory} \citep{chen2025factorychallenginghumanverifiedprompt} is a long-form factuality benchmark featuring human-verified challenging prompts. We adopt its hardest tier, Factory-Hard, which comprises 421 questions.

\textbf{FAVA} \citep{mishra2024finegrained} is a fine-grained hallucination benchmark consisting of 200 information-seeking queries that demand factual knowledge for accurate, multi-source long-form answers. Following \citet{DBLP:conf/nips/LinGOXLY024}, we use a subset of 151 prompts for evaluation.

\subsection{Baselines}
\label{sec:appendix_baselines}

We compare our method with two representative reinforcement learning approaches for improving long-form factuality: (i) RL with continuous VeriScore reward \citep{chen2025learningreasonfactuality}, and (ii) RL with Binary Retrieval-Augmented Reward (Binary RAR) \citep{chen2025traintruthskillsbinary}. Both methods optimize LRMs via on-policy RL, but differ in reward design.

\paragraph{RL Formulation}

RL for language models treats generation as a policy optimization problem. Given a prompt $x$, a language model $\pi_\theta$ generates a response $y \sim \pi_\theta(\cdot \mid x)$. The objective is to maximize a reward function $r(x, y)$ while constraining the policy to remain close to a reference model $\pi_{\text{ref}}$:
\begin{equation}
\label{eq:rl_objective_appendix}
\max_{\pi_\theta} \;
\mathbb{E}_{\substack{x \sim \mathcal{D} \\ y \sim \pi_\theta(\cdot \mid x)}}
\Big[
r(x, y)
- \beta \,
\mathbb{D}_{\mathrm{KL}}\!\big(
\pi_\theta(\cdot \mid x)
\;\|\;
\pi_{\mathrm{ref}}(\cdot \mid x)
\big)
\Big],
\end{equation}
where $\mathcal{D}$ is the prompt dataset and $\beta$ controls the KL regularization strength.

\paragraph{GRPO Optimization}

Both baselines optimize Equation \ref{eq:rl_objective_appendix} using GRPO \citep{shao2024deepseekmathpushinglimitsmathematical}. For each prompt $x$, a group of responses $\{y_i\}_{i=1}^n$ is sampled from the old policy $\pi_{\text{old}}$, and the policy $\pi_\theta$ is updated via:
{
\footnotesize
\begin{equation}
\label{eq:grpo_objective_appendix}
\begin{aligned}
\max_{\pi_\theta} \;
&
\mathbb{E}_{\substack{x \sim \mathcal{D}\\ \{y_i\}_{i=1}^{n} \sim \pi_{\text{old}}(\cdot|x)}}
\Bigg[
\frac{1}{n} \sum_{i=1}^{n} \frac{1}{|y_i|}
\sum_{t=1}^{|y_i|}
\\
&
\min\!\left(
\frac{\pi_\theta(y_i^t \mid y_i^{<t}, x)}
     {\pi_{\text{old}}(y_i^t \mid y_i^{<t}, x)} A_i,\;
\operatorname{clip}\!\left(
\frac{\pi_\theta(y_i^t \mid y_i^{<t}, x)}
     {\pi_{\text{old}}(y_i^t \mid y_i^{<t}, x)},
1-\epsilon,\,
1+\epsilon
\right) A_i
\right)
- \beta\,\mathbb{D}_{\mathrm{KL}}(\pi_\theta \,\|\, \pi_{\text{ref}})
\Bigg],
\end{aligned}
\end{equation}
}
where the advantage $A_i$ and KL regularization $\mathbb{D}_{\mathrm{KL}}$ are defined as:
\begin{equation}
A_i =
\frac{
r(x, y_i) - \operatorname{mean}[\,r(x, y_1), ..., r(x, y_n)\,]
}{
\operatorname{std}[\,r(x, y_1), ..., r(x, y_n)\,]
}
\end{equation}
\begin{equation}
\mathbb{D}_{\mathrm{KL}}(\pi_\theta \,\|\, \pi_{\text{ref}})
=
\frac{\pi_{\text{ref}}(y_i \mid x)}{\pi_\theta(y_i \mid x)}
-
\log\frac{\pi_{\text{ref}}(y_i \mid x)}{\pi_\theta(y_i \mid x)}
- 1
\end{equation}

\paragraph{RL with VeriScore Reward}

The first baseline employs a continuous factuality reward based on VeriScore. A generated response $y$ is decomposed into atomic claims $\mathcal{A} = \{a_1, \dots, a_{|\mathcal{A}|}\}$, each verified against retrieved evidence. The factual precision is defined as:
\begin{equation}
R_{\text{fact}}(y) =
\frac{1}{|\mathcal{A}|}
\sum_{i=1}^{|\mathcal{A}|}
\mathbb{I}\left(a_i \text{ is supported}\right).
\end{equation}

To avoid degenerate solutions (e.g., overly short responses), the reward further incorporates detail and relevance:
\begin{equation}
r_{\text{Veri}}(x, y) =
R_{\text{fact}}(y)
+ \lambda \log(1 + F(y))
+ \mu \cdot R_{\text{rel}}(x, y),
\end{equation}
where $F(y)$ denotes the number of supported claims, and $R_{\text{rel}}$ quantifies answer relevance as evaluated by an LLM acting as a judge.

\paragraph{RL with Binary Retrieval-Augmented Reward}

The second baseline adopts a binary reward based on retrieval-augmented verification. Given $(x, y)$, a set of documents $\mathcal{C}(x, y)$ is retrieved, and a verifier checks whether the response contradicts the evidence. The reward is defined as:
\begin{equation}
r_{\text{RAR}}(x, y) =
\begin{cases}
1, & \text{if no contradiction is found}, \\
0, & \text{otherwise}.
\end{cases}
\end{equation}

Compared to continuous rewards, Binary RAR provides a sparse but robust training signal that penalizes any factual inconsistency in the response.

\subsection{Metrics}
\label{sec:appendix_metrics}

We adopt VeriScore \citep{song-etal-2024-veriscore} as our primary evaluation metric. VeriScore assesses both the factuality and helpfulness of long-form generation through a three-stage pipeline: (i) claim extraction, (ii) evidence retrieval, and (iii) claim verification.
Given a response $y$, VeriScore first extracts a set of verifiable claims $\mathcal{A} = \{a_1, \dots, a_{|\mathcal{A}|}\}$. 
Each claim $a_i \in \mathcal{A}$ is then issued as a query to a search engine to retrieve supporting evidence $E_{a_i}$. 
The claim is subsequently verified against the retrieved evidence to determine whether it is supported.

An ideal response should exhibit both high factuality (i.e., minimal hallucination) and high helpfulness (i.e., sufficient coverage and completeness). To capture the trade-off between these two aspects, VeriScore computes an $F_1$ score based on precision and recall.

Factuality is defined as
\begin{equation}
P(y) = \frac{S(y)}{|\mathcal{A}|},
\end{equation}
where $S(y)$ denotes the number of supported claims and $|\mathcal{A}|$ is the total number of claims in the response.

Helpfulness is defined as
\begin{equation}
R(y) = \min\left(\frac{S(y)}{K}, 1\right),
\end{equation}
where $K$ represents the expected number of correct claims for the target domain. In practice, following the implementation of \citet{song-etal-2024-veriscore}, $K$ is set to the median number of supported claims extracted across all model responses within the target domain.

The final score is computed as the harmonic mean of factuality and helpfulness:
\begin{equation}
F_1 =
\begin{cases}
\frac{2 P(y)\,R(y)}{P(y) + R(y)}, & \text{if } S(y) > 0, \\
0, & \text{otherwise}.
\end{cases}
\end{equation}

We use GPT-5 as the backbone model for the VeriScore implementation.

\subsection{Data Curation}

\paragraph{Prompt Selection}

Collecting natural, high-quality, and diverse prompts that reflect realistic user interactions is crucial for effective model training. 
To this end, we build upon ELI5 \citep{fan-etal-2019-eli5}, a large-scale long-form question answering dataset comprising approximately 270K Reddit threads.
From this corpus, we aim to identify prompts that require long-form responses with verifiable factual content. 
We formulate this as a two-criteria filtering problem and implement it using a unified classification framework powered by GPT-5.

Specifically, we consider two criteria: 
(i) whether the prompt itself is factually grounded, and 
(ii) whether answering the prompt requires long-form generation with factual knowledge. 
Each criterion is operationalized via a dedicated classification prompt.
For the first criterion, we use a fact-checking prompt (Table \ref{tab:appendix_prompt_fact_checking}) to assess whether a prompt contains incorrect, impossible, or fictional information. 
For the second criterion, we use a factual knowledge requirement prompt (Table \ref{tab:appendix_prompt_knowledge_requirement}) to determine whether producing a high-quality answer requires both long-form generation and factual knowledge.
Applying this pipeline yields two high-quality subsets: 5K prompts for Calibration-Aware Structured Supervision and 3K prompts for Calibration-Aware Policy Distillation.

\paragraph{Answer Projection}

Given a prompt, its associated reasoning trajectory with reliability estimates, and the original final response, we introduce an answer projection procedure that transforms raw outputs into supervision targets aligned with both factual reliability and calibration-aware training objectives.

Our approach projects the final answer onto a subset of reasoning steps filtered by reliability signals. Specifically, we retain only those reasoning sentences labeled as \texttt{<reliable>} as authoritative sources of factual content, while permitting sentences marked as \texttt{<nonverifiable>} to contribute solely to discourse structure (e.g., coherence and fluency) without introducing new factual claims.

To operationalize the transformation, we design a constrained rewriting procedure (Table \ref{tab:appendix_prompt_answer_projection}) that revises the original answer under strict controls. This procedure enforces three principles:
(i) \textbf{faithfulness}, ensuring that all factual content is directly supported by reliable reasoning steps;
(ii) \textbf{conservativeness}, whereby information derived from unreliable reasoning is suppressed or excluded; and
(iii) \textbf{non-expansiveness}, preventing the introduction of information beyond the original reasoning trace.
The resulting projected answers provide high-quality supervision signals for downstream training, particularly in settings requiring fine-grained control over factual correctness and reliability.

\clearpage
\subsection{Prompt Template}

\begin{table}[!htbp]
\centering
\caption{Fact-checking prompt template.}
\label{tab:appendix_prompt_fact_checking}
\renewcommand{\arraystretch}{1.3}
\begin{tabular}{p{0.95\linewidth}}
\toprule

\texttt{You are a strict fact-checker.} \\ \\

\texttt{Given the following prompt:} \\ \\

\texttt{<Here is the prompt>} \\ \\

\texttt{Your task is to assess whether the prompt contains any factually incorrect, impossible, or fictional information. Please score the factual accuracy of the prompt using the scale below:} \\ \\

\texttt{0 = entirely fictional or impossible} \\
\texttt{1 = mostly incorrect or fabricated} \\
\texttt{2 = contains clear factual errors} \\
\texttt{3 = partially correct but includes questionable or uncertain claims} \\
\texttt{4 = mostly accurate with only minor uncertainty} \\
\texttt{5 = fully accurate with no detectable issues} \\ \\

\texttt{Return ONLY a JSON object in the following format:}
\begin{verbatim}
{
  "factual_score": <0-5>,
  "explanation": "short reason for the score"
}
\end{verbatim} \\

\bottomrule
\end{tabular}
\end{table}

\begin{table}[!htbp]
\centering
\caption{Factual knowledge requirement prompt template.}
\label{tab:appendix_prompt_knowledge_requirement}
\renewcommand{\arraystretch}{1.3}
\begin{tabular}{p{0.95\linewidth}}
\toprule

\texttt{Given the following prompt:} \\ \\

\texttt{<Here is the prompt>} \\ \\

\texttt{Your task is to determine whether answering this prompt correctly and in a high-quality manner REQUIRES both long-form generation and factual knowledge.} \\ \\

\texttt{Definition:} \\
\texttt{- Return **1** if producing a correct and high-quality answer requires both long-form generation and factual knowledge.} \\
\texttt{- Return **0** if the prompt can be adequately answered without long-form generation or without relying on factual knowledge.} \\ \\

\texttt{Return ONLY a JSON object in the following format:}
\begin{verbatim}
{
  "requires_factual_knowledge": 1 or 0,
  "explanation": "short reason for the judgment"
}
\end{verbatim} \\

\bottomrule
\end{tabular}
\end{table}

\begin{table}[!htbp]
\centering
\caption{Answer projection prompt template.}
\label{tab:appendix_prompt_answer_projection}
\renewcommand{\arraystretch}{1.3}
\begin{tabular}{p{0.95\linewidth}}
\toprule

\texttt{You are given a user question, a reasoning process (wrapped within <think></think> tags), and a final answer (wrapped within <answer></answer> tags) derived from that reasoning process. Each sentence in the reasoning process is annotated with a reliability\_score, which can be <unreliable>, <reliable>, or <nonverifiable>. A score of <nonverifiable> indicates that the sentence contains no verifiable factual content.} \\ \\

\texttt{Your task is to revise the final answer to improve factual accuracy by relying exclusively on reasoning sentences whose factuality\_score is <reliable>, while allowing sentences with a factuality\_score of <nonverifiable> to be used for non-factual, structural purposes only.} \\ \\

\texttt{Guidelines:} \\ \\

\texttt{* Reasoning sentences with a factuality\_score of <reliable> should be treated as fully correct and authoritative. The revised final answer should be based solely on these sentences, without hedging, reinterpretation, or additional verification.} \\

\texttt{* Downweight, soften, or omit information that depends on reasoning sentences with a factuality\_score of <unreliable>, even if such information appears relevant or plausible.} \\

\texttt{* Reasoning sentences with a factuality\_score of <nonverifiable> may be used to preserve the structure, flow, and coherence of the original answer, but must not serve as the basis for any factual assertions.} \\

\texttt{* Pay particular attention to potentially error-prone factual elements, including dates, names, numerical values, locations, and specific claims.} \\

\texttt{* Ensure the revised final answer remains coherent, fluent, and well-structured, and preserves the original answer’s overall structure and sentence ordering as much as possible, even if some content must be reduced or removed.} \\

\texttt{* Do NOT introduce any new information. The revised final answer must be fully and directly supported by the original reasoning process, without adding new assumptions, interpretations, or inferences.} \\ \\

\texttt{Output only the revised final answer, wrapped within <revised\_answer></revised\_answer> tags.} \\ \\

\texttt{User Question: <Here is the question>} \\

\texttt{Reasoning Process: <Here is the reasoning>} \\

\texttt{Final Answer: <Here is the response>} \\

\texttt{Revised Final Answer:} \\

\bottomrule
\end{tabular}
\end{table}

\clearpage
\section{Main Results}
\label{sec:appendix_main_results}

In this section, we provide a fine-grained analysis of model performance by decomposing VeriScore into its two components: factuality (precision) and helpfulness (recall). The detailed results are shown in Tables \ref{tab:appendix_main_results}, \ref{tab:appendix_main_results_precision} and \ref{tab:appendix_main_results_recall}.

\paragraph{CASS consistently improves factuality.}

Across all model families and scales, CASS leads to substantial gains in factuality. For example, Qwen3-8B improves from 60.24 to 70.84 (+10.6), and Llama-3.1-8B improves from 62.90 to 70.42 (+7.5). These gains are consistently larger than those achieved by RL-based baselines, indicating that calibration-aware generation is more effective at reducing hallucinations. This improvement aligns with the design of calibrated exploration and selective commitment, which explicitly suppress unreliable reasoning before it propagates to the final output.

\paragraph{Helpfulness exhibits a controlled trade-off.}

Compared to factuality, helpfulness shows a more nuanced pattern. CASS generally leads to a moderate decrease in recall, reflecting its conservative commitment strategy that filters out uncertain content. However, this reduction is relatively small compared to the substantial gains in factuality, resulting in improved overall VeriScore. Importantly, when combined with CAPD, recall is often partially recovered or even improved (e.g., Qwen3-4B and Llama-3.2-3B), suggesting that distillation helps the model better balance informativeness and reliability.

\paragraph{CAPD further improves the factuality-helpfulness balance.}

CAPD consistently enhances overall performance, particularly for smaller models. From a fine-grained perspective, CAPD often maintains or slightly improves factuality while recovering helpfulness, leading to better VeriScore. For instance, on Qwen3-4B, CAPD improves factuality (64.08 $\rightarrow$ 64.99) and significantly boosts helpfulness (68.71 $\rightarrow$ 73.09), demonstrating that on-policy distillation helps correct over-conservative behaviors introduced by CASS.

\paragraph{Overall: improved factuality with minimal loss in helpfulness.}

Taken together, the results demonstrate that CASS achieves a favorable trade-off between factuality and helpfulness. While it slightly reduces recall due to conservative commitment, it yields significantly larger gains in precision, leading to consistent improvements in VeriScore. With CAPD, the trade-off is further optimized, achieving both high factuality and strong helpfulness. These findings support our hypothesis that calibration-aware generation improves long-form factuality by selectively committing to reliable reasoning rather than globally suppressing outputs.

\begin{table}[!htbp]
\centering
\small
\caption{Main results on five long-form factuality benchmarks measured by VeriScore. CASS significantly improves performance over baselines across all models, and further gains are achieved with CAPD, demonstrating the effectiveness of calibration-aware generation.}
\label{tab:appendix_main_results}
\begin{tabular}{l*{3}{S[table-format=2.2, table-column-width=1.55cm]}*{3}{S[table-format=2.2, table-column-width=1.25cm]}}
\toprule
{\textbf{Models}} & {\textbf{AlpacaFact}} & {\textbf{Biography}} & {\textbf{FactBench}} & {\textbf{Factory}} & {\textbf{FAVA}} & {\textbf{AVG}} \\

\midrule

\textbf{Qwen3-4B} & 67.93 & 35.27 & 71.25 & 64.37 & 64.18 & 60.60 \\
+ Abstention & 46.00 & 23.40 & 45.92 & 25.91 & 48.13 & 37.87 \\
+ RL (VeriScore) & 68.29 & 32.00 & 72.35 & 61.58 & 65.45 & 59.93 \\
+ RL (Binary RAR) & 67.94 & 32.43 & 71.51 & 62.35 & 64.43 & 59.73 \\
\rowcolor{mygray} + CASS & 71.16 & 39.05 & 74.02 & 65.41 & 66.52 & 63.23 \\
\rowcolor{mygray} + CASS + CAPD & \textbf{76.44} & \textbf{40.62} & \textbf{75.38} & \textbf{67.30} & \textbf{70.76} & \textbf{66.10} \\

\midrule
\addlinespace

\textbf{Qwen3-8B} & 73.65 & 39.90 & 75.18 & 68.19 & 71.21 & 65.63 \\
+ Abstention & 33.70 & 21.93 & 28.37 & 22.68 & 44.50 & 30.24 \\
+ RL (VeriScore) & 74.09 & 40.04 & 73.80 & 65.72 & 72.13 & 65.16 \\
+ RL (Binary RAR) & 73.79 & 40.05 & 75.22 & 67.70 & 72.04 & 65.76 \\
\rowcolor{mygray} + CASS & 75.47 & \textbf{49.73} & 77.61 & 69.34 & 73.42 & 69.11 \\
\rowcolor{mygray} + CASS + CAPD & \textbf{78.24} & 49.64 & \textbf{79.79} & \textbf{70.37} & \textbf{75.13} & \textbf{70.63} \\

\midrule
\addlinespace

\textbf{Qwen3-14B} & 76.05 & 43.47 & 78.26 & 69.02 & 70.67 & 67.49 \\
+ Abstention & 55.61 & 32.23 & 52.58 & 21.42 & 59.23 & 44.21 \\
+ RL (VeriScore) & 77.82 & 47.66 & 79.68 & 67.25 & 73.35 & 69.15 \\
+ RL (Binary RAR) & 76.33 & 46.50 & 76.95 & 68.49 & 74.11 & 68.48 \\
\rowcolor{mygray} + CASS & \textbf{80.77} & \textbf{54.70} & \textbf{81.87} & \textbf{71.29} & \textbf{76.63} & \textbf{73.05} \\

\midrule
\addlinespace

\textbf{Llama-3.2-3B} & 65.56 & 37.57 & 66.48 & 50.75 & 58.55 & 55.78 \\
+ Abstention & 11.01 & 4.69 & 7.44 & 7.10 & 9.24 & 7.90 \\
+ RL (VeriScore) & 69.22 & 37.87 & 68.95 & 55.17 & 66.15 & 59.47 \\
+ RL (Binary RAR) & 70.32 & 38.58 & 70.45 & 54.22 & 66.81 & 60.08 \\
\rowcolor{mygray} + CASS & 71.81 & \textbf{46.85} & 70.47 & 63.22 & 69.55 & 64.38 \\
\rowcolor{mygray} + CASS + CAPD & \textbf{75.48} & 42.74 & \textbf{73.44} & \textbf{63.70} & \textbf{69.92} & \textbf{65.06} \\

\midrule
\addlinespace

\textbf{Llama-3.1-8B} & 71.55 & 48.18 & 69.03 & 54.22 & 64.59 & 61.51 \\
+ Abstention & 22.51 & 33.94 & 16.04 & 20.13 & 28.24 & 24.17 \\
+ RL (VeriScore) & 75.60 & 54.99 & 73.30 & 60.95 & 74.89 & 67.95 \\
+ RL (Binary RAR) & 76.16 & 55.66 & 73.52 & 61.01 & 74.15 & 68.10 \\
\rowcolor{mygray} + CASS & \textbf{77.05} & \textbf{61.10} & \textbf{77.52} & \textbf{66.24} & \textbf{77.03} & \textbf{71.79} \\

\bottomrule
\end{tabular}
\end{table}

\begin{table}[!htbp]
\centering
\small
\caption{Main results on five long-form factuality benchmarks measured by factuality (precision) under VeriScore.}
\label{tab:appendix_main_results_precision}
\begin{tabular}{l*{3}{S[table-format=2.2, table-column-width=1.55cm]}*{3}{S[table-format=2.2, table-column-width=1.25cm]}}
\toprule
{\textbf{Models}} & {\textbf{AlpacaFact}} & {\textbf{Biography}} & {\textbf{FactBench}} & {\textbf{Factory}} & {\textbf{FAVA}} & {\textbf{AVG}} \\

\midrule

\textbf{Qwen3-4B} & 67.97 & 26.34 & 69.72 & 56.67 & 54.75 & 55.09 \\
+ Abstention & 54.16 & 18.37 & 52.85 & 25.25 & 51.23 & 40.37 \\
+ RL (VeriScore) & 68.84 & 23.06 & 71.78 & 53.27 & 60.60 & 55.51 \\
+ RL (Binary RAR) & 67.73 & 23.23 & 71.21 & 53.22 & 59.19 & 54.92 \\
\rowcolor{mygray} + CASS & 76.25 & 32.28 & 78.46 & 64.49 & 68.93 & 64.08 \\
\rowcolor{mygray} + CASS + CAPD & \textbf{78.69} & \textbf{33.46} & \textbf{79.14} & \textbf{64.67} & \textbf{69.01} & \textbf{64.99} \\

\midrule
\addlinespace

\textbf{Qwen3-8B} & 72.69 & 30.68 & 72.04 & 61.97 & 63.80 & 60.24 \\
+ Abstention & 38.97 & 18.69 & 32.96 & 23.41 & 48.32 & 32.47 \\
+ RL (VeriScore) & 73.82 & 29.68 & 75.30 & 56.89 & 65.79 & 60.30 \\
+ RL (Binary RAR) & 74.63 & 29.64 & 76.18 & 67.56 & 65.41 & 62.68 \\
\rowcolor{mygray} + CASS & \textbf{81.17} & \textbf{45.57} & \textbf{83.29} & \textbf{68.54} & \textbf{75.63} & \textbf{70.84} \\
\rowcolor{mygray} + CASS + CAPD & 80.23 & 43.97 & 82.90 & 67.45 & 73.93 & 69.70 \\

\midrule
\addlinespace

\textbf{Qwen3-14B} & 75.86 & 35.76 & 76.60 & 60.98 & 64.42 & 62.72 \\
+ Abstention & 65.56 & 25.90 & 60.15 & 21.12 & 63.69 & 47.28 \\
+ RL (VeriScore) & 79.29 & 36.52 & 80.89 & 59.14 & 68.88 & 64.94 \\
+ RL (Binary RAR) & 76.73 & 35.76 & 77.81 & 59.95 & 68.95 & 63.84 \\
\rowcolor{mygray} + CASS & \textbf{83.75} & \textbf{48.36} & \textbf{86.09} & \textbf{69.12} & \textbf{74.91} & \textbf{72.45} \\

\midrule
\addlinespace

\textbf{Llama-3.2-3B} & 67.43 & 31.98 & 67.90 & 55.31 & 59.33 & 56.39 \\
+ Abstention & 13.37 & 4.16 & 8.57 & 8.62 & 12.28 & 9.40 \\
+ RL (VeriScore) & 69.90 & 28.14 & 67.36 & 45.99 & 60.76 & 54.43 \\
+ RL (Binary RAR) & 71.13 & 28.94 & 68.45 & 44.64 & 59.79 & 54.59 \\
\rowcolor{mygray} + CASS & \textbf{78.37} & \textbf{43.57} & 74.12 & 61.41 & \textbf{69.21} & \textbf{65.34} \\
\rowcolor{mygray} + CASS + CAPD & 77.94 & 34.90 & \textbf{77.50} & \textbf{62.44} & 67.47 & 64.05 \\

\midrule
\addlinespace

\textbf{Llama-3.1-8B} & 75.91 & 41.36 & 71.27 & 58.93 & 67.05 & 62.90 \\
+ Abstention & 27.78 & 33.11 & 18.18 & 21.94 & 32.49 & 26.70 \\
+ RL (VeriScore) & 77.40 & 44.20 & 73.20 & 50.14 & 70.17 & 63.02 \\
+ RL (Binary RAR) & 75.75 & 44.34 & 73.66 & 50.83 & 68.48 & 62.61 \\
\rowcolor{mygray} + CASS & \textbf{80.05} & \textbf{52.21} & \textbf{82.02} & \textbf{61.60} & \textbf{76.23} & \textbf{70.42} \\

\bottomrule
\end{tabular}
\end{table}

\begin{table}[!htbp]
\centering
\small
\caption{Main results on five long-form factuality benchmarks measured by helpfulness (recall) under VeriScore.}
\label{tab:appendix_main_results_recall}
\begin{tabular}{l*{3}{S[table-format=2.2, table-column-width=1.55cm]}*{3}{S[table-format=2.2, table-column-width=1.25cm]}}
\toprule
{\textbf{Models}} & {\textbf{AlpacaFact}} & {\textbf{Biography}} & {\textbf{FactBench}} & {\textbf{Factory}} & {\textbf{FAVA}} & {\textbf{AVG}} \\

\midrule

\textbf{Qwen3-4B} & 73.02 & 61.63 & 79.63 & 81.59 & 78.96 & 74.97 \\
+ Abstention & 45.31 & 34.04 & 46.01 & 30.29 & 50.33 & 41.20 \\
+ RL (VeriScore) & 73.86 & 56.83 & 78.78 & 81.84 & 75.75 & 73.41 \\
+ RL (Binary RAR) & 73.27 & 58.00 & 76.81 & 82.78 & 74.91 & 73.15 \\
\rowcolor{mygray} + CASS & 72.16 & 53.15 & 75.94 & 72.53 & 69.77 & 68.71 \\
\rowcolor{mygray} + CASS + CAPD & 79.11 & 55.16 & 78.34 & 75.69 & 77.16 & 73.09 \\

\midrule
\addlinespace

\textbf{Qwen3-8B} & 79.64 & 65.18 & 84.23 & 82.66 & 85.50 & 79.44 \\
+ Abstention & 32.91 & 29.27 & 27.80 & 25.18 & 45.55 & 32.14 \\
+ RL (VeriScore) & 78.51 & 65.73 & 78.08 & 84.34 & 83.52 & 78.04 \\
+ RL (Binary RAR) & 77.62 & 66.12 & 79.75 & 73.14 & 84.47 & 76.22 \\
\rowcolor{mygray} + CASS & 76.24 & 57.92 & 78.59 & 76.85 & 75.79 & 73.08 \\
\rowcolor{mygray} + CASS + CAPD & 80.21 & 60.94 & 82.06 & 80.21 & 80.38 & 76.76 \\

\midrule
\addlinespace

\textbf{Qwen3-14B} & 80.68 & 64.40 & 84.99 & 86.06 & 83.30 & 79.89 \\
+ Abstention & 53.71 & 44.65 & 52.16 & 24.39 & 61.37 & 47.26 \\
+ RL (VeriScore) & 80.24 & 74.24 & 83.89 & 85.30 & 81.75 & 81.08 \\
+ RL (Binary RAR) & 79.79 & 71.27 & 81.09 & 87.37 & 83.74 & 80.65 \\
\rowcolor{mygray} + CASS & 82.38 & 67.14 & 83.60 & 79.62 & 81.46 & 78.84 \\

\midrule
\addlinespace

\textbf{Llama-3.2-3B} & 69.07 & 51.52 & 72.37 & 55.99 & 67.18 & 63.23 \\
+ Abstention & 10.87 & 5.46 & 7.57 & 7.09 & 8.54 & 7.91 \\
+ RL (VeriScore) & 73.96 & 63.00 & 75.68 & 77.95 & 76.75 & 73.47 \\
+ RL (Binary RAR) & 75.20 & 62.14 & 78.48 & 77.84 & 80.43 & 74.82 \\
\rowcolor{mygray} + CASS & 72.28 & 54.14 & 73.26 & 71.83 & 74.25 & 69.15 \\
\rowcolor{mygray} + CASS + CAPD & 78.24 & 58.40 & 76.03 & 71.80 & 76.62 & 72.22 \\

\midrule
\addlinespace

\textbf{Llama-3.1-8B} & 73.76 & 60.66 & 74.04 & 59.77 & 70.27 & 67.70 \\
+ Abstention & 21.69 & 38.02 & 16.28 & 22.21 & 28.55 & 25.35 \\
+ RL (VeriScore) & 79.40 & 76.03 & 78.45 & 84.40 & 84.77 & 80.61 \\
+ RL (Binary RAR) & 80.53 & 78.61 & 78.64 & 82.90 & 84.40 & 81.02 \\
\rowcolor{mygray} + CASS & 78.26 & 76.97 & 79.57 & 77.32 & 82.78 & 78.98 \\

\bottomrule
\end{tabular}
\end{table}

\clearpage
\section{Ablation on the Bucketing Threshold}
\label{sec:appendix_threshold_ablation}

In Calibration-Aware Generation (CAG), continuous factuality scores are discretized into reliability labels via a threshold-based bucketing function (Eq. \ref{equ:bucketing_general} and \ref{equ:bucketing_binary}). The threshold $\tau$ determines which reasoning steps are considered reliable, and therefore directly affects selective commitment in the final answer. In this section, we analyze how the bucketing threshold $\tau$ affects calibration-aware generation.

\paragraph{Experimental Setup}

We vary the threshold $\tau \in \{0.2, 0.3, 0.4, 0.5, 0.6\}$ and evaluate performance on three representative benchmarks: AlpacaFact, Biography, and FAVA. Experiments are conducted on two models, Llama-3.1-8B and Qwen3-14B. Performance is measured using VeriScore, which jointly evaluates factuality and helpfulness.

\paragraph{Results}

\begin{figure}[!tbp]
\centering
\includegraphics[width=\linewidth]{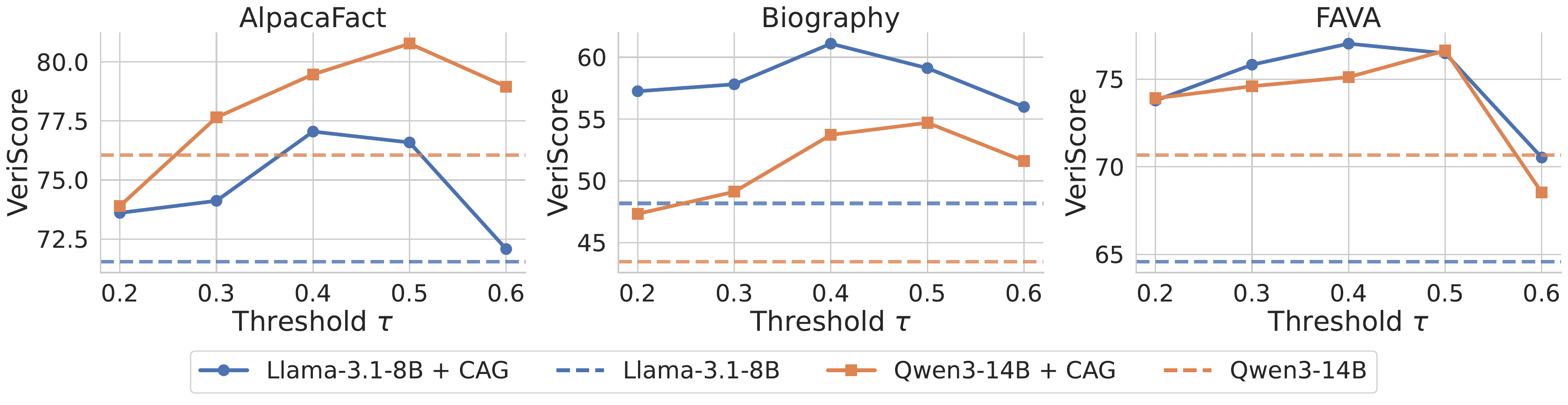}
\caption{Ablation on the bucketing threshold $\tau$. Performance is measured by VeriScore across different datasets and model families. Moderate thresholds achieve the best performance, reflecting a factuality--helpfulness trade-off: lower thresholds improve coverage but introduce hallucinations, while higher thresholds increase factuality at the cost of discarding useful information. The optimal threshold range is stable across datasets and models, and CAG consistently outperforms baselines across most threshold values, demonstrating robustness to threshold selection.}
\label{fig:appendix_threshold_ablation}
\end{figure}

Figure \ref{fig:appendix_threshold_ablation} shows the performance trends across different thresholds. We observe a consistent non-monotonic relationship between $\tau$ and performance across all datasets and models. In particular, moderate thresholds (around $\tau = 0.4$--$0.5$) achieve the best results, while both lower and higher thresholds lead to performance degradation.

When $\tau$ is small (e.g., $\tau = 0.2$), the model considers a large number of reasoning steps as reliable, including those with low factuality scores. As a result, selective commitment becomes less effective, and unreliable information is more likely to be propagated into the final answer, leading to decreased factuality. Conversely, when $\tau$ is large (e.g., $\tau = 0.6$), only highly confident steps are retained. While this improves precision, it also removes moderately reliable reasoning, resulting in overly conservative outputs with reduced informational coverage and lower helpfulness.

Importantly, CAG consistently outperforms the baseline models across most threshold values, indicating that calibration-aware generation is robust to moderate variations in threshold selection, while still benefiting from proper calibration of reliability signals.

\paragraph{Discussion}

The observed behavior reflects a fundamental factuality-helpfulness trade-off induced by thresholding. Lower thresholds favor helpfulness by retaining more reasoning steps but risk introducing hallucinations, whereas higher thresholds favor factuality at the cost of discarding useful information. The optimal threshold balances the two factors, aligning with the decision-theoretic interpretation of thresholding discussed in Appendix \ref{sec:appendix_bayes}.

Notably, the optimal range of $\tau$ is stable across datasets and model scales, suggesting that a single global threshold (e.g., $\tau = 0.4$) is sufficient in practice without per-task tuning.

\section{Limitations}
\label{sec:limitations}

While the proposed CAG framework demonstrates consistent improvements in long-form factuality across multiple models and benchmarks, several limitations remain.

\paragraph{Trade-offs between informativeness and conservativeness}

By prioritizing high-reliability content, CAG may introduce a slight bias toward more conservative outputs. However, empirical results indicate that any reduction in informativeness is minimal within the evaluated settings. Furthermore, CAG can be combined with complementary methods to further enhance informativeness while maintaining strong factual reliability.

\paragraph{Effects of model scale and capacity}

We observe that smaller models tend to benefit more from CAPD, indicating that accurate reliability estimation is more challenging for lower-capacity models. The extent to which calibration-aware generation scales with increasingly large models warrants further investigation.

\paragraph{Interaction with complementary techniques}

The proposed framework is designed to be method-agnostic and is readily compatible with approaches such as factuality-driven optimization and RAG. While we demonstrate its effectiveness across several representative settings, further investigation into its interaction with a broader range of training and inference techniques may yield additional insights.

\section{Societal Impacts and Ethical Considerations}
\label{sec:ethics}

Our work presents minimal potential for negative societal impact, primarily due to the use of publicly available datasets and models. This accessibility inherently reduces the risk of adverse effects on individuals or society.

\clearpage
\newpage
\section*{NeurIPS Paper Checklist}

\begin{enumerate}

\item {\bf Claims}
    \item[] Question: Do the main claims made in the abstract and introduction accurately reflect the paper's contributions and scope?
    \item[] Answer: \answerYes{} 
    \item[] Justification: The main claims presented in the abstract and introduction accurately reflect the paper’s contributions and scope (see Section \ref{sec:intro}).

\item {\bf Limitations}
    \item[] Question: Does the paper discuss the limitations of the work performed by the authors?
    \item[] Answer: \answerYes{} 
    \item[] Justification: See Appendix \ref{sec:limitations}. 

\item {\bf Theory assumptions and proofs}
    \item[] Question: For each theoretical result, does the paper provide the full set of assumptions and a complete (and correct) proof?
    \item[] Answer: \answerNA{} 
    \item[] Justification: The paper does not include theoretical results.

    \item {\bf Experimental result reproducibility}
    \item[] Question: Does the paper fully disclose all the information needed to reproduce the main experimental results of the paper to the extent that it affects the main claims and/or conclusions of the paper (regardless of whether the code and data are provided or not)?
    \item[] Answer: \answerYes{} 
    \item[] Justification: See Appendix \ref{sec:appendix_implementation}.

\item {\bf Open access to data and code}
    \item[] Question: Does the paper provide open access to the data and code, with sufficient instructions to faithfully reproduce the main experimental results, as described in supplemental material?
    \item[] Answer: \answerNo{} 
    \item[] Justification: We provide detailed descriptions of the implementation, including models, datasets, and training configurations (Appendix \ref{sec:appendix_implementation}). We will release the code and data to facilitate reproducibility upon acceptance.

\item {\bf Experimental setting/details}
    \item[] Question: Does the paper specify all the training and test details (e.g., data splits, hyperparameters, how they were chosen, type of optimizer) necessary to understand the results?
    \item[] Answer: \answerYes{} 
    \item[] Justification: See Appendix \ref{sec:appendix_implementation}.

\item {\bf Experiment statistical significance}
    \item[] Question: Does the paper report error bars suitably and correctly defined or other appropriate information about the statistical significance of the experiments?
    \item[] Answer: \answerYes{} 
    \item[] Justification: The paper reports statistical variability for key analyses. In particular, Figure \ref{fig:analysis_calibration_intervention} presents results with 95\% confidence intervals, reflecting variability across samples.

\item {\bf Experiments compute resources}
    \item[] Question: For each experiment, does the paper provide sufficient information on the computer resources (type of compute workers, memory, time of execution) needed to reproduce the experiments?
    \item[] Answer: \answerYes{} 
    \item[] Justification: See Appendix \ref{sec:appendix_implementation}.
    
\item {\bf Code of ethics}
    \item[] Question: Does the research conducted in the paper conform, in every respect, with the NeurIPS Code of Ethics \url{https://neurips.cc/public/EthicsGuidelines}?
    \item[] Answer: \answerYes{} 
    \item[] Justification: We ensure that all aspects of the research comply with the NeurIPS Code of Ethics. In particular, we strictly preserve anonymity throughout the submission and adhere to all relevant ethical guidelines outlined by NeurIPS.

\item {\bf Broader impacts}
    \item[] Question: Does the paper discuss both potential positive societal impacts and negative societal impacts of the work performed?
    \item[] Answer: \answerYes{} 
    \item[] Justification: See Appendix \ref{sec:ethics}.
    
\item {\bf Safeguards}
    \item[] Question: Does the paper describe safeguards that have been put in place for responsible release of data or models that have a high risk for misuse (e.g., pre-trained language models, image generators, or scraped datasets)?
    \item[] Answer: \answerNA{} 
    \item[] Justification: The paper poses no such risks.

\item {\bf Licenses for existing assets}
    \item[] Question: Are the creators or original owners of assets (e.g., code, data, models), used in the paper, properly credited and are the license and terms of use explicitly mentioned and properly respected?
    \item[] Answer: \answerYes{} 
    \item[] Justification: We properly cite all original code, models, and prior work, and adhere to all applicable licenses and usage terms.

\item {\bf New assets}
    \item[] Question: Are new assets introduced in the paper well documented and is the documentation provided alongside the assets?
    \item[] Answer: \answerNA{} 
    \item[] Justification: The paper does not release new assets.

\item {\bf Crowdsourcing and research with human subjects}
    \item[] Question: For crowdsourcing experiments and research with human subjects, does the paper include the full text of instructions given to participants and screenshots, if applicable, as well as details about compensation (if any)? 
    \item[] Answer: \answerNA{} 
    \item[] Justification: The paper does not involve crowdsourcing nor research with human subjects.

\item {\bf Institutional review board (IRB) approvals or equivalent for research with human subjects}
    \item[] Question: Does the paper describe potential risks incurred by study participants, whether such risks were disclosed to the subjects, and whether Institutional Review Board (IRB) approvals (or an equivalent approval/review based on the requirements of your country or institution) were obtained?
    \item[] Answer: \answerNA{} 
    \item[] Justification: The paper does not involve crowdsourcing nor research with human subjects.

\item {\bf Declaration of LLM usage}
    \item[] Question: Does the paper describe the usage of LLMs if it is an important, original, or non-standard component of the core methods in this research? Note that if the LLM is used only for writing, editing, or formatting purposes and does \emph{not} impact the core methodology, scientific rigor, or originality of the research, declaration is not required.
    \item[] Answer: \answerNA{} 
    \item[] Justification: The core method development in this research does not involve LLMs as any important, original, or non-standard components.

\end{enumerate}

\end{document}